\documentclass[10pt,twocolumn,letterpaper]{article}

\usepackage{cvpr}              
\usepackage{multicol}
\usepackage{multirow}
\usepackage{diagbox}

\usepackage[T1]{fontenc}
\usepackage[dvipsnames]{xcolor}

\definecolor{cvprblue}{rgb}{0.21,0.49,0.74}
\usepackage[pagebackref,breaklinks,colorlinks,citecolor=cvprblue]{hyperref}

\def\paperID{5547} 
\def\confName{CVPR}
\def\confYear{2024}

\title{Human Motion Prediction under Unexpected Perturbation}

\author{Jiangbei Yue\footnotemark[1]\\
\and
Baiyi Li\footnotemark[1]\\
\and
Julien Pettr\'{e}\footnotemark[2] \\
\and
Armin Seyfried\footnotemark[3]\\
\and
He Wang\footnotemark[4]
}

\begin{document}
\maketitle
\renewcommand{\thefootnote}{\fnsymbol{footnote}}
\footnotetext[1]{University of Leeds, United Kingdom}
\footnotetext[2]{INRIA Rennes, France}
\footnotetext[3]{Forschungszentrum J\"{u}lich, Germany}
\footnotetext[4]{corresponding author, he\_wang@ucl.ac.uk, University College London, United Kingdom}
\renewcommand{\thefootnote}{\arabic{footnote}}
\begin{abstract}
We investigate a new task in human motion prediction, which is predicting motions under unexpected physical perturbation potentially involving multiple people. Compared with existing research, this task involves predicting less controlled, unpremeditated and pure reactive motions in response to external impact and how such motions can propagate through people. It brings new challenges such as data scarcity and predicting complex interactions. To this end, we propose a new method capitalizing differential physics and deep neural networks, leading to an explicit Latent Differential Physics (LDP) model. Through experiments, we demonstrate that LDP has high data efficiency, outstanding prediction accuracy, strong generalizability and good explainability. Since there is no similar research, a comprehensive comparison with 11 adapted baselines from several relevant domains is conducted, showing LDP outperforming existing research both quantitatively and qualitatively, improving prediction accuracy by as much as 70\%, and demonstrating significantly stronger generalization.
\end{abstract}    
\section{Introduction}
\label{sec:intro}

Human motion prediction aims to predict the future movements given the past motions, which has been heavily studied in computer vision~\cite{xu2023eqmotion, guo2023back, peng2023trajectory, xu2023joint, xu2022stochastic}. Deviating from existing research, we are interested in a new task setting: predicting human motions, on both individual and group levels, under unexpected physical perturbation. On the individual level, physical perturbation causes reactive motions as opposed to active motions. On the group level, such perturbations can propagate through people while possibly being intensified, \eg a push at the back of a line of people could be transferred all the way to the front. These motions have not been investigated. Incorporating physical perturbation potentially extends motion prediction to new application domains \eg balance recovery in biomechanics~\cite{hsiao2008biomechanical, brodie2018optimizing}, reactive motions for character animation~\cite{arikan2005pushing, geijtenbeek2012interactive}, crowd crush induced by pushing~\cite{wang2019modeling, chen2023extended}, humanoid robots~\cite{katic2003survey, kahraman2020fuzzy}, \etc.

Incorporating physical perturbation in prediction imposes new challenges. First, the motions are purely reactive and less controlled such that they are less smooth and less coordinated among body parts. Furthermore, this perturbation can propagate through people when they are packed and the space to recover balance is restricted, such that an attempt to recover balance relies on pushing others. Last but not least, unlike existing research, the data for motion prediction under perturbation is extremely scarce. Not only is it rare to capture full-body motions under such circumstances, but it is also difficult to record the interactions between people, \eg forces of pushes.

Before deep learning, many areas have formulated this problem, which can be broadly divided into two categories. The first is physics-based where human bodies are simplified into connected rigid bodies~\cite{liu2010sampling, wang2019modeling}. The reaction to push is solved via optimization to compute what forces are needed to recover balance~\cite{ott2011posture, mordatch2012discovery}, or through carefully tuning feed-forward controllers~\cite{liu2010sampling, liu2015improving}. These methods, despite aiming to mimic the balance recovery of humans, do not learn from human data and therefore cannot predict human motions. Alternatively, reactions to perturbation can be learned from data via regression~\cite{wei2011physically}, optimization~\cite{xie2021physics}, or reinforcement learning~\cite{won2020scalable}. Comparatively, this type of method tends to generate more human-like motions, but they are not designed for prediction. 

Recently, deep learning~\cite{xu2023eqmotion, xu2023joint, peng2023trajectory,wang2019spatio} have dominated human motion prediction, but they cannot be adapted for our problem. First, most datasets only contain single-body motions without external perturbation. Even when multiple people are captured, it is not under unexpected perturbation. To predict push propagation, one would still need to measure information \eg contact forces between people, ground friction, muscle forces, \etc, which are all absent. This data scarcity essentially rules out most deep-learning methods. Furthermore, there is also little work in modelling the physical/bio-mechanical interactions that can potentially propagate through people. Current research includes motion forecasting, generation and synthesis. Most motion forecasting methods~\cite{li2020dynamic, guo2023back, xu2023eqmotion} are for a single person, with a few recent exceptions~\cite{xu2022stochastic, peng2023trajectory, xu2023joint} but not involving perturbation. Alternatively, our problem could be formulated as motion generation conditioned on external perturbation. However, current methods~\cite{tang2022real,tang2023rsmt, petrovich2021action, tevet2022human, zhang2023remodiffuse,chen2020dynamic} again do not explicitly model close interactions among multiple people caused by perturbation. Theoretically, motion synthesis~\cite{ling2020character, won2022physics, wang2014energy,wang2013harmonic} is a possibility, which potentially can predict motions under perturbation. But they require dense control signals to guide the synthesis, or/and extensive physical simulation. Therefore, it requires manual labor or/and is difficult to scale to many people.      

To address the aforementioned challenges, we need a model that has  \textit{high data efficiency}, \textit{strong generalizability} and can model \textit{interactions between people}. In other words, this model needs to be able to learn from a small number of samples, can predict accurately in situations similar to the data, and is capable of generating plausible motions in drastically different scenarios. To this end, we propose a new deep-learning model for human motion prediction under unexpected perturbation. To address the data scarcity, we propose a scalable differentiable physics (DP) model for the human body, to learn the balance strategy and interaction propagation between people, inspired by recent DP research~\cite{gong2022fine, wang2021sim2sim, yue2023human}. However, naively following existing DP approaches means we would need to make the full-body simulation differentiable for each individual. Not only is motion intrinsically indifferentiable due to \eg foot contact, but full-body physical models are too computationally expensive to scale. Therefore, we propose a latent DP space where the full-body physics is reduced into a differentiable inverted pendulum model (IPM)~\cite{olfati2001nonlinear,kajita20013d, kwon2017momentum, hwang2018real}, and the full-body poses are mapped to and recovered from the IPM. At the low level, the IPM governs body physics and learns key forces such as ground friction and balance recovery. As the IPM is simple, the required data is small. At the high level, we use neural networks to recover the full-body pose from the IPM, which also does not require much data as the IPM provides strong guidance. We refer to our model as the Latent Differentiable Physics (LDP) model. Note different from other latent physics models where the dimensionality reduction is implicit~\cite{shen2021high, wiewel2019latent}, ours is explicit and physically meaningful (\ie mapping from full-body to IPM).

We show LDP can learn from very limited data and perform well under many widely used metrics. Since there is no similar work to our best knowledge, we adapt a wide range of baseline methods in the most relevant areas (motion forecasting, motion generation and motion synthesis), in single-person and multi-people scenarios, for comparison. The results demonstrate that LDP outperforms them both quantitatively and qualitatively. Notably, our model exhibits remarkable generalizability. It can accommodate unseen out-of-distribution perturbations, group sizes, and group formations, potentially extending our research beyond human motion prediction into broader areas, \eg crowd simulation. Furthermore, owing to the explicit physics model, our model possesses a distinctive feature: explainability, providing plausible explanations for the predicted motion.  Formally, our contributions include:
\begin{itemize}
    \item A new task: human motion prediction under unexpected perturbation. To our best knowledge, this is the first deep-learning paper addressing this problem. 
    \item A novel differentiable physics model in human motion prediction that explicitly considers physical interactions.
    \item A new differentiable IPM model that learns body physics under complex interactions. 
    \item A novel differentiable interaction model that can learn interactions and interaction propagation.
\end{itemize}

\section{Related Work}
\label{sec:rw}





\begin{figure*}[tb]
  \centering
   \includegraphics[width=1\textwidth]{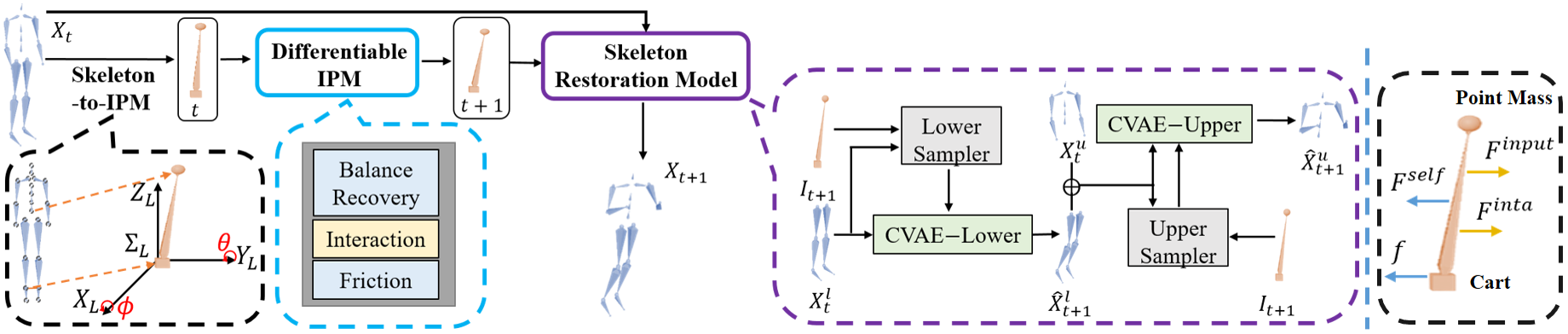}
   \caption{\textbf{Overview of our model.} Given a frame $X_t$, it is first mapped into the IPM space via Skeleton-to-IPM to get its IPM state $I_t$. Then $I_t$ is simulated for one step via Differentiable IPM to compute $I_{t+1}$. Lastly, the full-body frame $X_{t+1}$ is recovered from $I_{t+1}$ via Skeleton Restoration Model. The IPM is shown in the right figure.  The full-body state $X$ is represented by joint positions.}
   \label{fig:overview}
\end{figure*}
\textbf{Human Motion Prediction.} Compared with traditional statistical machine learning~\cite{wang2005gaussian,lehrmann2014efficient}, deep learning has dominated human motion prediction recently. It can be formulated as a sequence-to-sequence task modelled by Recurrent Neural Networks~\cite{fragkiadaki2015recurrent, jain2016structural, pavllo2018quaternet}. Also, human skeletons can be seen as graphs so that spatio-temporal graph convolutions can be employed~\cite{li2020dynamic, cui2021towards, dang2021msr, li2021multiscale, zang2021few}. Transformer-based methods~\cite{aksan2021spatio, cai2020learning} use the attention mechanism to capture spatial and temporal correlations. Recently, there has been a surge of interest in multi-people motion prediction~\cite{wang2021multi, xu2023joint, peng2023trajectory, xu2022stochastic}.  MRT~\cite{wang2021multi} models the social interactions between humans via a global encoder. JRFormer~\cite{xu2023joint} exploits the joint relation representation for modelling the interactions where physical interactions are considered implicitly. However, existing methods share a common limitation - they do not consider unexpected perturbations, restricting their applications in predicting actively planned/controlled motions. Additionally, explicit physical interactions between people have often been overlooked in these methods. Our model extends the research to a more challenging scenario involving unexpected perturbation and perturbation propagation. The explicit physics knowledge in our model enables it to achieve better prediction, generalizability, and explainability.

\textbf{Traditional Research on Balance Recovery} Relevant research has been conducted in other fields where traditional methods mainly focus on modelling balance recovery strategies in response to perturbations~\cite{hsiao2008biomechanical, brodie2018optimizing, katic2003survey, kahraman2020fuzzy, ott2011posture, chen2023extended}. Brodie \etal~\cite{brodie2018optimizing} analyzed the biomechanical mechanisms in the balance recovery following an unexpected perturbation such as trips and slips. Chen \etal~\cite{chen2023extended} studied the dynamics of individuals under pushing in crowds. A new controller was proposed to recover balance for bipedal robots under perturbation~\cite{ott2011posture}. In parallel, some traditional methods aim to synthesize reactive motions to perturbation~\cite{arikan2005pushing, mordatch2012discovery,wei2011physically,peng2018deepmimic,luo2023perpetual}. Arikan \etal\cite{arikan2005pushing} proposed an algorithm for selecting and adjusting the motions from data to synthesize the motion for animating virtual characters being pushed. \cite{peng2018deepmimic,luo2023perpetual} explored how to turn the given motions under perturbation into physically valid ones. Overall, traditional methods cannot predict motions under perturbation, either because they do not learn from data or have limited learning capacity. By contrast, we incorporate DP with deep neural networks to predict such human motions.

\textbf{Differentiable Physics.} DP is an emerging field focusing on combining traditional physics models with deep learning techniques, to provide high data efficiency and explainability. Consequently, many domains have investigated differentiable physics such as robotics~\cite{Werling2021fast, le2023differentiable, chen2022imitation}, physics~\cite{IOLLO200187, JARNY19912911}, computer vision~\cite{yue2023human,yue2022human},  and computer graphics~\cite{NEURIPS2019_28f0b864, gong2022fine}. We propose the first explicit latent differentiable physics model for human motion prediction under unexpected physical perturbation.

\section{Methodology}
\label{sec:method}




\textbf{Problem Definition.} Given a motion with multiple people, we denote the skeletal pose of the $n$th person at frame $t$ as $X_t^n\in\mathbb{R}^{J\times3}$ where $J$ is the joint number. Unlike existing research aiming to predict $p$ frames $\{\hat{X}_{T-p+1:T}^n \}_{n=1}^{N}$ given $k$ frames $\{X_{1:k}^n \}_{n=1}^{N}$ history, we minimize the required history due to limited data. Given the initial frame $\{ X_0^n \}_{n=1}^{N}$ and the input forces $F^{input}$, we aim to predict the following T frames, by solving an initial problem:
\begin{equation}
\resizebox{.9\hsize}{!}{ 
    $\{\hat{X}_{1:T}^n \}_{n=1}^{N} =\mathcal{S}_{\gamma}(\{ X_0^n \}_{n=1}^{N}, IPM_{\eta}(\mathcal{M}(\{ X_0^n \}_{n=1}^{N}), F^{input}))$
}
\label{eq:overview}
\end{equation}
where $\{\hat{X}_{1:T}^n \}_{n=1}^{N}$ is the predicted $T$ frames. $\mathcal{M}$ is a Skeleton-to-IPM mapping $\mathcal{M}: \mathcal{X}\rightarrow\mathcal{I}$ where $\mathcal{X}$ and $\mathcal{I}$ are the state space of skeleton poses (represented by joint positions) and the IPM respectively. $IPM_{\eta}$ is a Differentiable IPM with learnable parameters $\eta$. Finally, $\mathcal{S}_{\gamma}$ is the inverse mapping, \ie Skeleton Restoration Model, $\mathcal{S}_{\gamma}: \mathcal{I}\rightarrow\mathcal{X}$, reconstructing full-body skeleton pose from IPM states, with learnable parameters $\gamma$. An overview of our model is shown in \cref{fig:overview}. Given a motion, we map the full-body poses into their corresponding states of an IPM~\cite{kajita20013d, kwon2017momentum, hwang2018real} as $\{I_0^n\}_{n=1}^{N} = \mathcal{M}(\{ X_0^n \}_{n=1}^{N})$. By simulating the IPM forward in time via $IPM_{\eta}$, it can learn the key parameters $\eta$. The interaction forces between people are also learned simultaneously. Meanwhile, our Skeleton Restoration Model $\mathcal{S}_{\gamma}$ recovers the full-body poses from the predicted IPM states from $IPM_{\eta}$. For training, we minimize the mean squared error (MSE) between the predicted $\{ \hat{X}_{1:T}^n \}_{n=1}^{N}$ and the ground-truth poses $\{X_{1:T}^n \}_{n=1}^{N}$: 
\begin{align}
    Loss = MSE(\{ \hat{X}_{1:T}^n \}_{n=1}^{N}, \{ X_{1:T}^n \}_{n=1}^{N})
    \label{eq:loss}
\end{align}
where we need to specify $\mathcal{S}_{\gamma}$, $IPM_{\eta}$ and $\mathcal{M}$ in \cref{eq:overview}. We give key equations and model information below and refer the readers to the supplementary material (SM) for details.

\subsection{Latent Physics Space for Full-body Motions}
\label{subsec:ipm}
\subsubsection{Background and Skeleton-to-IPM Mapping}
We first introduce $IPM_{\eta}$ and $\mathcal{M}$ in \cref{eq:overview}. Differentiable physics (DP) has shown extremely high data efficiency because physics can act as a strong inductive bias and eliminates the reliance on large amounts of training data~\cite{ding2021dynamic, wang2021sim2sim, yue2023human}. For our model, a key design choice is to choose a DP model that has the right level of granularity while being scalable. Among many possible choices from full-body physics~\cite{al2012trajectory} to simple rods~\cite{xiang2010predictive}, we choose the Inverted Pendulum Model (IPM)~\cite{kajita20013d, kwon2017momentum, hwang2018real} as it can fully capture balance loss and recovery while being scalable.



Our IPM has a massless rod mounted to a cart with a point mass at the end of the rod (\cref{fig:overview} right). Denoting its state $\mathcal{I}\ni I_t = [x_t,y_t,\theta_t,\phi_t]\in\mathbb{R}^4$ at time step $t$ where $[x,y]$ is the coordinates of the cart in the xy-plane and $[\theta, \phi]$ is the rotation angles of the rod around $Y_L$ axis and $X_L$ axis in the local coordinate system $\Sigma_L$, respectively. Our full-body pose $X$ is represented by 22 joint positions. $\mathcal{M}$ in \cref{eq:overview} is defined as (\cref{fig:overview} left): the hip joint is mapped onto the point mass, and the midpoint between the two ankle joints is mapped onto the center of the cart. The point mass and the cart jointly determine the two angles $[\theta, \phi]$. 

Next, $IPM_{\eta}$ is defined. Given the initial IPM state, we can simulate it in time by solving \cref{eq:ipm_mt} repeatedly~\cite{olfati2001nonlinear}:
\begin{equation}
  M(I_t, l_{t})\Ddot{I}_t + C(I_t, \Dot{I}_t, l_{t}) + G(I_t, l_{t}) = F^{net}_t
  \label{eq:ipm_mt}
\end{equation}
where $M\in\mathbb{R}^{4\times4}$, $C\in\mathbb{R}^{4\times1}$ and $G\in\mathbb{R}^{4\times1}$ are the inertia matrix, the Centrifugal/Coriolis matrix, and the external force such as gravity, which are all functions of state $I_t$, its first-order derivative $\Dot{I}_t$ and the rod length $l_{t}$. While the standard IPM has a fixed rod length, we allow it to change as the distance between the hip and the middle of two ankles can drastically change in human motions. Therefore, we also predict $l_{t}$ at each time step. Overall given the net force $F^{net}_t\in\mathbb{R}^4$ and the rod length $l_{t}$, we can solve \cref{eq:ipm_mt} for the next state $I_{t+1}$ via a semi-implicit scheme $\Dot{I}_{t+1} = \Dot{I}_t + \triangle t \Ddot{I}_t$ and $I_{t+1} = I_t + \triangle t \Dot{I}_{t+1}$, where $\triangle t$ is the time step. 

Finally, the learnable parameters $\eta$ in $IPM_{\eta}$ parameterize $F_t^{net}$ and the rod length $l_{t}$, where the formulation differs between single-person and multi-people, and will be elaborated later. It's notable that $F_t^{net}$ in \cref{eq:ipm_mt} is the generalized force. Using the generalized force (instead of the Euler force) keeps the motion equation simple, and its entries have explicit physical meanings as shown later.

\subsubsection{Single-Person Prediction via Differentiable IPM}
\label{subsubsec:single}


Under single-person, we only consider Balance-Recovery and Friction (blue blocks in \cref{fig:overview} Differentiable IPM) when predicting $F^{net}$. Specifically, we consider three forces:
\begin{equation}
    F_t^{net} = F_t^{self} + f_t + F_t^{input}
\label{eq:netf_sin}
\end{equation}
where $F_t^{self}$, $f_t$, and $F_t^{input}$ are the balance recovery force, the ground friction and the external perturbation. The Balance-Recovery module learns $F_t^{self}$ which is further decomposed into $F_t^{self} = F_t^{self-pd} + F_t^{self-nn}$. This decomposition is because $F_t^{self}$ is the muscle force at the hinge of the rod which serves two purposes. The first one is to give a feed-forward torque $F_t^{self-pd}$ to react to perturbation for balance recovery, and the second is to give a torque correction $F_t^{self-nn}$ for tracking observed motions. In generalized forces, we parameterize $F_t^{self-pd}$ by proportional derivative (PD) control:
\begin{align}
    F_t^{self-pd} = K_pe_t + K_d\Dot{e}_t \text{, }e_t = s_d - s_t
\end{align}
where $e_t$ is the PD state error, $K_p$ and $K_d$ are the control parameters. Different from the IPM state, the current PD state is $s_t=[\Dot{x}_t, \Dot{y}_t, \theta_t, \phi_t]$ and the desired PD state $s_d$ is $[0, 0, 0, 0]$.   
In other words, we assume people tend to recover to the upright body pose and zero linear velocity after unexpected perturbation, which is a widely accepted assumption~\cite{kuo1995optimal, stodolka2016balance, li2012gyroscopic}. However, $F_t^{self-pd}$ only captures the general balance recovery strategy. To mimic the data, we parameterize $F_t^{self-nn}$ with a Long Short Term Memory (LSTM) network:
\begin{equation}
    F_t^{self-nn} = LSTM([\theta_t, \phi_t, \Dot{x}_t, \Dot{y}_t, \Dot{\theta}_t, \Dot{\phi}_t, M]),
    \label{eq:lstm}
\end{equation}
where $M$ is the mass of the person.  

Ground friction $f_t$ is the main reason for successful self-balance and therefore needs to be explicitly considered. In generalized forces, friction affects the IPM motion via damping~\cite{olfati2001nonlinear}. So we parameterize $f_t = -\mu[\Dot{x}_t, \Dot{y}_t, 0, 0]$, where the parameter $\mu$ is a learnable positive scalar and shared by all people for simplicity. The damping force only directly influences the cart motion.

Finally, to compute \cref{eq:ipm_mt}, we also need to predict the change of the rod length $l_t$, where we employ a multi-layer perception (MLP):
\begin{align}
    \label{eq:rod}
    \triangle l_t = MLP([\theta_t, \phi_t, \Dot{x}_t, \Dot{y}_t, \Dot{\theta}_t, \Dot{\phi}_t, F_t^{self}, M, l_t])
\end{align}
where $l_t$ is the rod length at time step $t$. We predict the rod length at the next time step by $l_{t+1} = l_t + \triangle l_t$. Finally, after obtaining the prediction of $F_t^{net}$ and $l_{t}$ at every time step $t$, we can calculate the next IPM state by solving \cref{eq:ipm_mt} via the semi-implicit scheme mentioned above.

\subsubsection{Multi-people with Differentiable Interaction}
\label{subsubsec:multi}
When there is more than one person, the complexity increases quickly. The main reason is that the interaction propagation among people is: (1) complex, \eg complicated contact positions/duration/forces. (2) hard to capture in data. Therefore, we propose to consider them as latent variables that cannot be directly observed. But again large amounts of data would be needed if we only relied on data to infer these variables. Therefore we model the interactions in the reduced IPM space, rather than the original space, so that it becomes a Differential Interaction Model (DIM).

Our DIM models a differentiable interaction force between any two IPMs and is learned in the Interaction module (the yellow block in \cref{fig:overview} Differentiable IPM). The overall net force on an IPM in multi-people then becomes:
\begin{align}
    \label{eq:netf_mul}
    F_t^{net}& = F_t^{self} + F_{t,n}^{inta} + f_t + F_t^{input}    
\end{align}
where $F_t^{self}$, $f_t$ and $F_t^{input}$ are the same as \cref{eq:netf_sin}. Note all forces are learned and shared among all people, so that we can generalize to an arbitrary number of people later. $F_{t, n}^{inta}\in\mathbb{R}^4$ is the new interaction force:
\begin{equation}
\resizebox{.9\hsize}{!}{
$F_{t,n}^{inta} = \sum_{j \in \Omega_{t,n}} F_{t,nj}^{inta} = \sum_{j \in \Omega_{t,n}} F_{t,nj}^{inta-bs} + F_{t,nj}^{inta-nn}$
}
    \label{eq:inta}
\end{equation}
where $\Omega_{t, n}$ is the neighborhood of the person $n$ at time $t$. $F_{t,nj}^{inta}$ is the interaction force applied onto person $n$ from her/his neighbor $j\in\Omega_{t, n}$. We model two factors in $F_{t,nj}^{inta}$: $F_{t,nj}^{inta-bs}$ and $F_{t,nj}^{inta-nn}$. The first $F_{t,nj}^{inta-bs}$ represents a consistent and trackable repulsive tendency when two IPMs get close, while $F_{t,nj}^{inta-nn}$ captures the variations of the repulsion. So we expect $F_{t,nj}^{inta-bs}$ to capture most of the interaction while $F_{t,nj}^{inta-nn}$ being a supplement. To this end, we separate the dimensions of an IPM state $I = [x,y,\theta,\phi]$ into two groups $[x,y]$ and $[\theta,\phi]$ and treat them separately as $F_{nj}^{bs-xy}\in\mathbb{R}^2$ and $F_{nj}^{bs-\theta \phi}\in\mathbb{R}^2$, such that $F_{t,nj}^{inta-bs} = [F_{nj}^{bs-xy}, F_{nj}^{bs-\theta \phi}]^{\mathbf{T}}$, where we omit the time subscript $t$ and the superscript $inta$ for simplicity. 



For $F_{nj}^{bs-xy}$, we define a repulsive potential energy between two close IPMs which leads to a repulsive force: 
\begin{align}
    \label{eq:inta_bsxy}
    F_{nj}^{bs-xy}(r_{nj}) = - \nabla_{r_{nj}} \mathcal{U}[b(r_{nj})], \ \  \mathcal{U}[b] = ue^{-\frac{b} {\sigma}} \\
    b = \frac{1}{2}\sqrt{(\lVert r_{nj} \rVert +  \lVert r_{nj} - \triangle t \Dot{r}_{jn} \rVert)^2 - \lVert \triangle t \Dot{r}_{jn} \rVert ^ 2 }.
\end{align}
where $r_{nj} = r_n - r_j$ is the relative position of the carts of a person and his/her neighbor j, \ie $r_n$ is the vector $[x,y]$ in the IPM state $I_n$. The $\mathcal{U}[b]$ is the repulsive potential with elliptical equipotential lines, and $u$ and $\sigma$ are hyper-parameters. $b$ is the semi-minor axis of the ellipse where $\Dot{r}_{jn} = \Dot{r}_j - \Dot{r}_n$ is the relative velocity. 

For $F_{nj}^{bs-\theta \phi}$, we treat it as a force with a constant magnitude (tunable hyperparameter) and apply it on $\theta$ and $\phi$ independently. Although the magnitude is constant, its directions can vary in different situations. We explain it for $\theta$ and the same principle applies to $\phi$. On the high level, we need to decide the direction of $F_{nj}^{bs-\theta \phi}$ based on the states of two close IPMs. $\theta$ can be positive, zero and negative. For two IPMs, this produces a total of 9 possible states, which we detail in the SM.


After defining $F_{t,nj}^{inta-bs}$, we explain $F_{t,nj}^{inta-nn}$ which should capture the variation of interactions. Unlike $F_{t,nj}^{inta-bs}$ where we can define an explicit form, we learn $F_{t,nj}^{inta-nn}$ via an MLP:
\begin{equation}
\resizebox{.85\hsize}{!}{ 
    $F_{nj}^{nn} = MLP([x_{nj}, y_{nj}, \theta_n, \phi_n, \theta_j, \phi_j, \Dot{x}_{nj}, \Dot{y}_{nj}, \Dot{\theta}_{nj}, \Dot{\phi}_{nj}])$
    }
\label{eq:interaction}
\end{equation}
where $x_{nj} = x_n - x_j$ and $\Dot{x}_{nj} = \Dot{x}_{n} - \Dot{x}_{j}$. $y_{nj}$, $\Dot{y}_{nj}$, $\Dot{\theta}_{nj}$ and $\Dot{\phi}_{nj}$ are computed in a similar fashion.




\subsection{Skeleton Restoration Model}
\label{subsec:skel}
To predict full-body motion, we recover the full-body pose from the predicted IPM states. This is divided into two steps as shown in \cref{fig:overview}. We first recover the lower body from the IPM state, then recover the upper body from both the IPM state and the recovered low body. There are two reasons for this design. First, the Skeleton-to-IPM mapping dictates that the IPM has a higher correlation with the lower body than with the upper body. Also, the dynamics of the lower body and the upper body are relatively independent~\cite{tang2022real, kormushev2011upper}, \ie similar low-body motions can correspond to different upper-body motions, \eg different styles in walking. Therefore, we use two models to recover the lower body and the upper body, respectively.
Overall, although the Skeleton Restoration Model involves deep neural networks, the required data is small as there is strong IPM guidance. 



\textbf{Lower Body Restoration.} We use a Conditional Variational Autoencoder (CVAE)~\cite{petrovich2021action, won2022physics, tang2022real} (CVAE-Lower in \cref{fig:overview}) to learn a Normal distribution of the lower body ${X}_{t+1}^l$ in the latent space conditioned on ${X}_{t}^l$. During inference, since ${X}_{t+1}^l$ is unavailable, we train a sampler (Lower Sampler) to sample the latent space to generate the next frame ${\hat{X}}_{t+1}^l$.  The Lower Sampler network is an MLP. It takes as input ${X}_{t}^l$, $I_{t+1}$, and outputs a latent code of CVAE-Lower which is then decoded. Overall, CVAE-Lower takes as input the current lower body ${X}_{t}^l$ and the predicted IPM state $I_{t+1}$, to predict the next lower body ${\hat{X}}_{t+1}^l$,  essentially reconstructing the lower body under the IPM guidance.

\textbf{Upper Body Restoration.} Similarly, we also use a CVAE named CVAE-Upper, except this time we use both the lower body predicted by CVAE-Lower ${\hat{X}}_{t+1}^l$ and the current upper body ${X}_{t}^u$ as the condition. A sampler (Upper Sampler) is also used to take as input $I_{t+1}$, ${\hat{X}}_{t+1}^l$ and ${X}_{t}^u$, and sample the latent space of CVAE-Upper, which is then decoded to predict the upper body at the next frame ${\hat{X}}_{t+1}^u$.

\subsection{Training with Auxiliary Losses}
In summary, the learnable parameters of our model include: the LSTM (\cref{eq:lstm}), the MLPs (\cref{eq:rod}, \cref{eq:interaction}), the ground friction coefficent $\mu$, CVAE-Lower, CVAE-Upper, Lower Sampler and Upper Sampler. Other than the main loss in \cref{eq:loss}, we also use other auxiliary losses such as foot sliding, IPM state MSE, \etc We also pre-train some components for initialization. Due to space limit, all details including training/prediction algorithms, implementation details, hyperparameters, \etc. are in the SM.

\section{Experiments}
\label{sec:exps}
\subsection{Dataset and Metrics}
\begin{figure}[tb]
  \centering
   \includegraphics[width=0.7\linewidth]{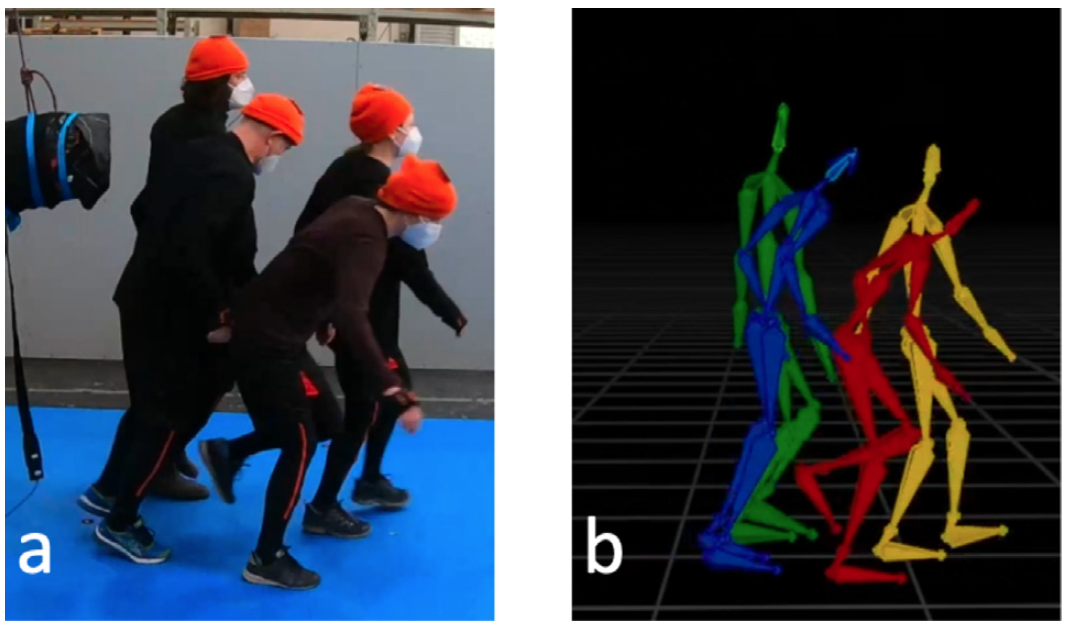}
   \caption{FZJ Push~\cite{feldmann2023forward}. The blue agent was pushed by the punch bag and then he pushed other people.}
   \label{fig:data}
\end{figure}
\begin{figure*}[tb]
  \centering
   \includegraphics[width=1.0\linewidth]
   {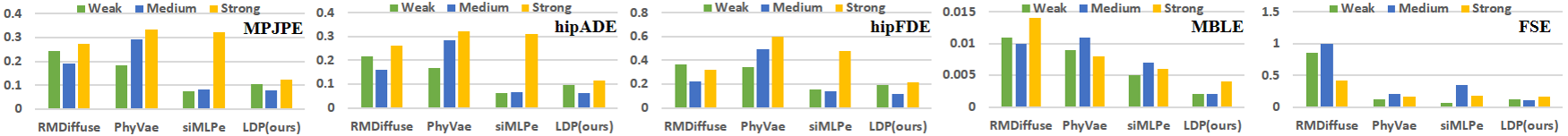}
    \includegraphics[width=1.0\linewidth]{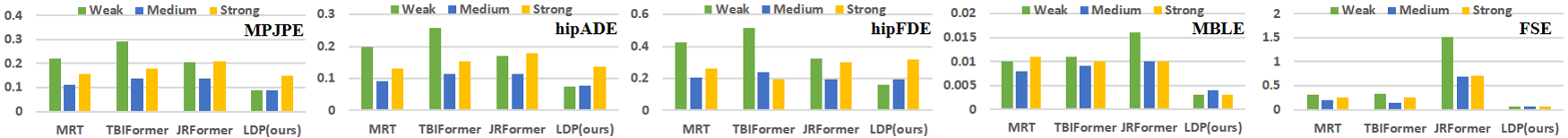}
   \caption{Perturbations with different magnitudes in single-person (top) and multi-people (bottom).}
   \label{fig:inten}
\end{figure*}
Data for our problem is extremely scarce compared with other human motion prediction research. The only publicly available dataset, to our best knowledge, is a new dataset ~\cite{feldmann2023forward} named FZJ Push. The dataset includes standing individuals, groups of four, and groups of five, with one person pushed by a punching bag unexpectedly and the push is propagated through the group. In total, the dataset includes only 45 single-person motions and 63 multi-person motions. This is considerably less than data normally used for human motion prediction. As shown later, the necessity of a model with high data efficiency is crucial. The motion is recorded at 60 Hz. Shown in \cref{fig:data} a, a hanging punch bag is operated by a person to give pushes of various magnitudes to one person in the group. Then the skeletal motions (\cref{fig:data} b) are recorded. There is a pressure sensor measuring the pushing forces on the punching bag. However, the pushing forces between people are not recorded. We discard redundant data such as frames in waiting. 

For evaluation, we adopt five widely used metrics~\cite{xu2023eqmotion, xu2022stochastic, tang2022real}: Mean Per Joint Position Error (MPJPE) in meters, Average Displacement Error at the hip (hipADE) in meters, Final Displacement Error at the hip (hipFDE) in meters, Mean Bone Length Error (MBLE) in meters, and Foot Skating Error (FSE) in centimeters. Details and justifications for these metrics are in the SM.

\begin{table}[tb]
\footnotesize
  \centering
  \begin{tabular}{p{1.4cm} p{1cm} p{0.9cm} p{0.9cm} p{0.8cm} p{0.8cm}}
    \toprule
    Method & MPJPE & hipADE & hipFDE & MBLE & FSE \\
    \midrule
    A2M &0.403 & 0.386 & 0.730 & 0.019 & 0.200  \\
    \midrule
    ACTOR &0.362 & 0.338 & 0.591 & 0.020 & 0.434  \\
    \midrule
    MDM &0.500 & 0.424 & 0.686 & 0 & 2.567  \\
    \midrule
    RMDiffuse & 0.228 & 0.202 & 0.299 & 0.011 & 0.790  \\
    \midrule
    PhyVae & 0.260 & 0.249 & 0.460 & 0.009 & 0.170 \\
    \midrule
    siMLPe & 0.130 & 0.117 & 0.226 & 0.006 & 0.182 \\
    \midrule
    EqMotion & 0.296 & 0.270 & 0.543 & 0.064 & 1.552 \\
    \midrule
    Ours & 0.097 & 0.086 & 0.171 & 0.002 & 0.131 \\
    \bottomrule

    \toprule
    MRT & 0.162 & 0.140 & 0.282 & 0.010 & 0.256  \\
    \midrule
    DuMMF & 0.312 & 0.285 & 0.480 & 0 & 3.194  \\
    \midrule
    TBIFormer &0.204 & 0.177 & 0.305 & 0.010 & 0.234  \\
    \midrule
     JRFormer & 0.181 & 0.152 & 0.260 & 0.012 & 0.932 \\
    \midrule
    Ours & 0.106 & 0.092 & 0.218 & 0.003 & 0.069 \\
    \bottomrule
  \end{tabular}
  \caption{Metrics in single-person (top) and multi-people (bottom).}
  \label{tab:metrics}
\end{table}
\subsection{Baselines}
There is no similar work in human motion prediction to our best knowledge, so we carefully review a wide spectrum of research in motion prediction, synthesis and generation, and choose the latest methods in each field for comparison. Specifically, we choose 11 baselines: A2M~\cite{guo2020action2motion}, ACTOR~\cite{petrovich2021action}, MDM~\cite{tevet2022human}, RMDiffuse~\cite{zhang2023remodiffuse}, PhyVae~\cite{won2022physics}, siMLPe~\cite{guo2023back} and EqMotion~\cite{xu2023eqmotion} for the single-person scenario, and MRT~\cite{wang2021multi}, DuMMF~\cite{xu2022stochastic}, TBIFormer~\cite{peng2023trajectory} and JRFormer~\cite{xu2023joint} for the multi-person scenario. The specific adaptation varies according to the baseline, and we give the details in the SM. One notable difference is our model only requires the first frame with the perturbation force during inference, while the other methods tend to require much more information such as multiple frames.

\subsection{Quantitative Results}


The single-person comparison is shown in \cref{tab:metrics} top. Despite requiring the minimal information, our model still achieves the best performance on all metrics except the MBLE. MDM obtained 0 MBLE because its parameterization is joint angle based, \ie no bone-length change incurred. A joint angle parameterization could also work with our model but in practice, we find a joint-position-based parameterization works better. Across different metrics, LDP outperforms the best baseline by as much as 25.38\%, 26.50\%, 24.34\%, 66.67\%, and 22.94\% on MPJPE, hipADE, hipFDE, MBLE, and FSE respectively, excluding the MBLE of MDM. We tend to attribute the higher performance to the explicit physics-based inductive biases embedded in the design of LDP. Furthermore, we look into performances under perturbations with different magnitudes (weak, medium and strong) in \cref{fig:inten} top, where we only include the best three baselines and leave the full comparison in SM. Stronger pushes lead to stronger responses and tend to be harder to predict. This is especially obvious in metrics related to motion tracking, \ie MPJPE, hipADE and hipFDE, where as the push becomes stronger, the errors become larger. Comparatively, LDP consistently outperforms other baselines, demonstrating its effectiveness in strong perturbations. In addition, compared with weak and medium pushes, LDP has a slower error increment under strong pushes, in contrast to the more volatile performances of other baselines, showing better generalizability. Overall, LDP either ranks as the best or is close to the top performance across metrics and perturbation levels.

The results under the multi-people scenario are shown in \cref{tab:metrics} bottom. The MBLE of DuMMF is 0 because it employs joint-angle-based parameterization. Multi-people is a challenging task for all methods. On all metrics, LDP outperforms all  baselines by at least 
34.57\%, 34.29\%, 16.15\%, 70\%, and 70.51\% on MPJPE, hipADE, hipFDE, MBLE, and FSE, respectively, (excluding the MBLE of DuMMF). Again note the baselines get much more input information compared with our model. Moreover, we show detailed analysis under perturbations with different magnitudes in \cref{fig:inten} bottom, with the three best baselines. One challenge in multi-people is to predict the onset and duration of interactions. The baseline methods need to learn the interactions by purely fitting the data, while our method learns them as a latent physical process. Consequently, none of the baselines can predict well, \eg they predict moving without being pushed or not moving while being pushed, while our model can learn to predict the interactions and their propagation well. Overall, our model achieves or is close to the best performance across metrics and perturbation levels.

\subsection{Qualitative Results}
\begin{figure}[tb]
  \centering
   \includegraphics[width=1.0\linewidth]{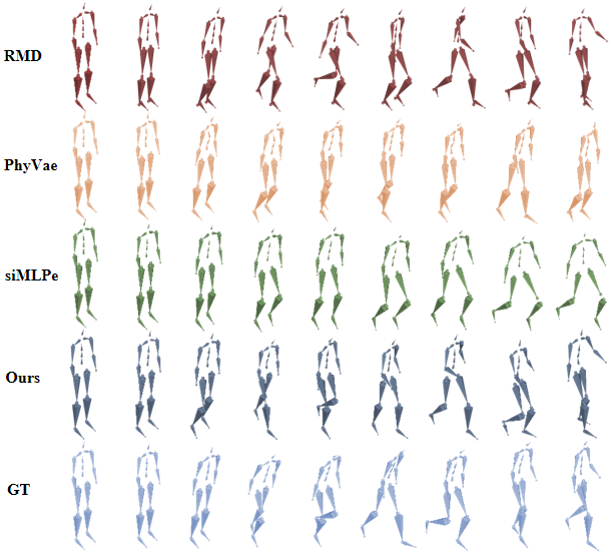}
   \caption{Visual Results in the Single-person scenario.}
   \label{fig:quality_s}
\end{figure}
\begin{figure*}[tb]
  \centering
   \includegraphics[width=1\linewidth]{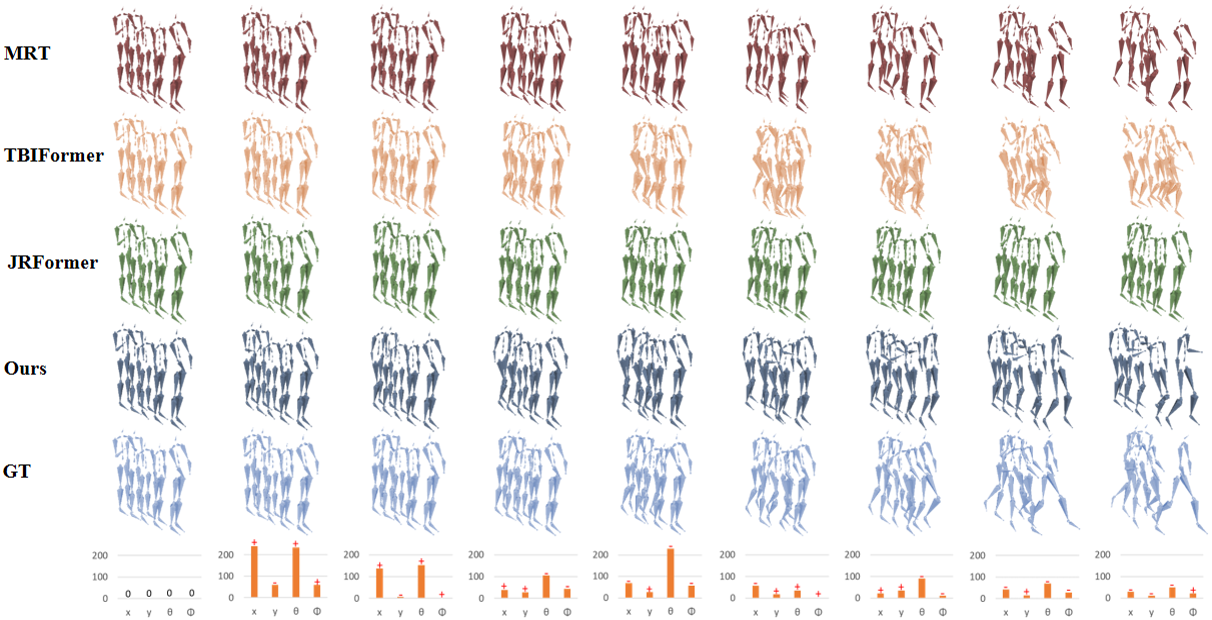}
   \caption{Multi-people comparison. The last row shows the learned net force on the second (from the left) person. The bar height indicates the magnitude and the sign indicates the direction, where the people move in the positive direction of the x-axis.}
   \label{fig:quality_M}
\end{figure*}
We visually compare our methods with the best three baselines under single-person in \cref{fig:quality_s}. Our prediction has the highest quality and is the most similar to the ground truth. RMDiffuse severely violates bone lengths, especially around ankles, and generates jittering motions.  PhyVae predicts walking but with rather small steps. siMLPe predicts only a single step. The multi-people scenario is much harder (\cref{fig:quality_M}), where both individual reactions and interactions need to be predicted. MRT and TBIFormer suffer from serious intersections between individuals. JRFormer predicts merely subtle movements that deviate considerably from the ground truth. Our model generates the most similar prediction to the ground truth.

\textbf{Explainability} In \cref{fig:quality_M} bottom, we show the learned net forces on the second person (from left), to provide plausible explanations of the predicted motion. This person remains still initially under zero net force, then experiences a push from the first person, resulting in forces in $x$ and $\theta$, and small forces in $y$ and $\phi$. Then the third person is pushed by the second, resulting in the change of the net force on the second person from positive to negative in $x$ and $\theta$. Finally, the second person recovers the balance. Our model predicts the motion results from plausible forces, and therefore possess strong explainability.

\subsection{Generalization}
\begin{figure*}[tb]
  \centering
   \includegraphics[width=0.9\linewidth]{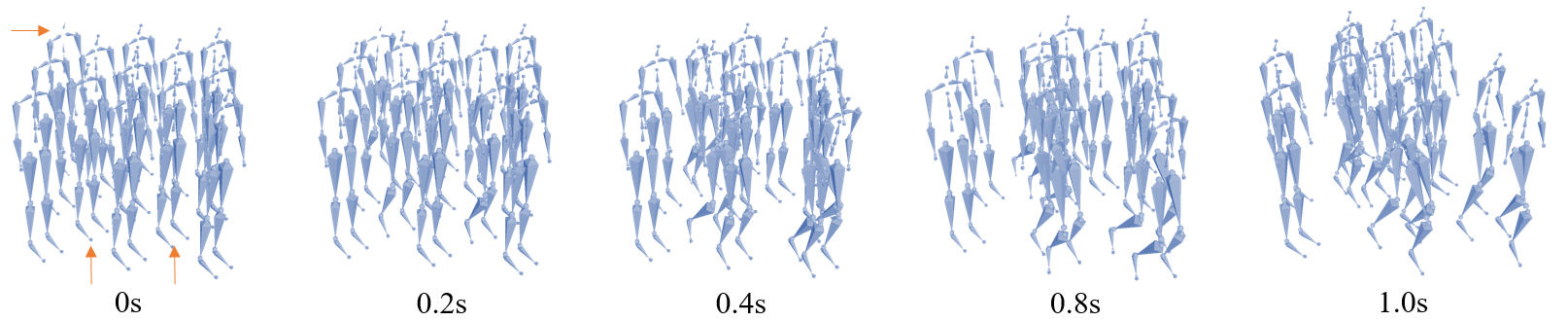}
   \caption{A 13-person group in a diamond formation with three people (indicated by orange arrows) being pushed.}
   \label{fig:quality_gen}
\end{figure*}

LDP can easily generalize to out-of-distribution scenarios, \eg unseen pushes, more people, different formations, \etc. Since there is no ground truth, we show the visual result of a challenging generalization scenario in \cref{fig:quality_gen}, where 13 people stand in a diamond formation and 3 of them indicated by the orange arrows are pushed. Note the data only contain up to 5 people in simple formations such as one or two lines. So this 13-people formation is totally out-of-distribution. However, our model can still generate plausible motions for the entire group, given only the initial poses and the perturbation forces, demonstrating strong generalizability. More experiments can be found in the SM.    

\subsection{Ablation Study}
\begin{table}[tb]
\scriptsize
  \centering
  \begin{tabular}{p{1.8cm} p{0.8cm} p{0.8cm} p{0.8cm} p{0.8cm} p{0.8cm}}
    \toprule
   Method & MPJPE & hipADE & hipFDE & MBLE & FSE \\
    \midrule
    no IPM, Full &0.217 & 0.195 & 0.341 & 0.007 & 0.196  \\
    \midrule
    no IPM, Low-up& 0.206 & 0.184 & 0.320 & 0.009 & 0.313 \\
    \midrule
    IPM, Full & 0.110 & 0.094& 0.242 & 0.004 & 0.126 \\
    \midrule
    IPM, Low-up & 0.106 & 0.092 & 0.218 & 0.003 & 0.069 \\
    \bottomrule
  \end{tabular}
  \caption{Ablation study with (1) IPM and no IPM, (2) Full body and Lower-up body pose reconstruction.}
  \label{tab:ablation}
\end{table}
The Differentiable IPM and the Skeleton Restoration Model are two key components of our model. We conduct the ablation study to assess the effectiveness of them. There are four combinations: with/without IPM, and full-body restoration or separate restoration (first lower body then upper body). When the IPM is absent, the next frame is directly predicted by either one full-body CVAE (Full) or two CVAEs with one for the lower body and the other for the upper body (Low-up). Without IPM, there are also no samplers (Lower Sampler and Upper Sampler in \cref{fig:overview}) so we need to directly sample in the latent space of the CVAEs. We randomly sample the latent space 3 times when predicting the next frame and average the results. In contrast, with IPM, we can train the samplers to only sample once to predict the next frame.

Results are shown in \cref{tab:ablation}. When there is no IPM, the performance deteriorates significantly across all metrics. With the IPM guidance, all metrics are significantly improved. Further, the Low-up separation of the body improves the performance further across all metrics under the IPM guidance, especially on the FSE. However, it exhibits limited effectiveness without the IPM guidance, even resulting in a bad FSE. This is because IPM states have strong correlations with the lower body, without which the Low-up is unable to improve the performance significantly even when the lower body is separately predicted.







\section{Conclusion}
We proposed a new task, human motion prediction under unexpected perturbation, which extends human motion prediction into new application domains. To this end, we have identified and overcome new challenges \eg data scarcity and interaction modelling, by proposing a new class of deep learning models based on differentiable physics. Our model outperforms existing methods despite requiring far less information and shows strong generalization to unseen scenarios. One limitation is our method requires explicit modelling of the physical process, making the model not as general as black-box deep neural nets that can be plug-and-play on data. However, we argue this is mainly driven by the data scarcity. Also, it brings stronger generalizability and interpretability. In future, we will investigate more general physics models that can potentially accommodate more diversified physical interactions between people.

A big difference between other existing datasets~\cite{ionescu2013human3,von2018recovering} and the dataset FZJ Push is the former is active motions while the latter is passive balance recovery balance. Therefore, the self-actuating force is very different in these two types of motions. We will also explore LDP on action motions in future.   

\section*{Acknowledgements}
The project received funding from the European Union’s Horizon 2020 research and innovation programme under grant agreement No 899739 CrowdDNA.
\clearpage
\setcounter{page}{1}
\setcounter{section}{0}
\maketitlesupplementary


Code and pre-processed data are available: https://github.com/realcrane/Human-Motion-Prediction-under-Unexpected-Perturbation.

\section{Additional Experiments}
\subsection{Single-person Results}
\subsubsection{More Comparison}
\begin{figure*}[tb]
    \centering
    \includegraphics[width=1.0\linewidth]{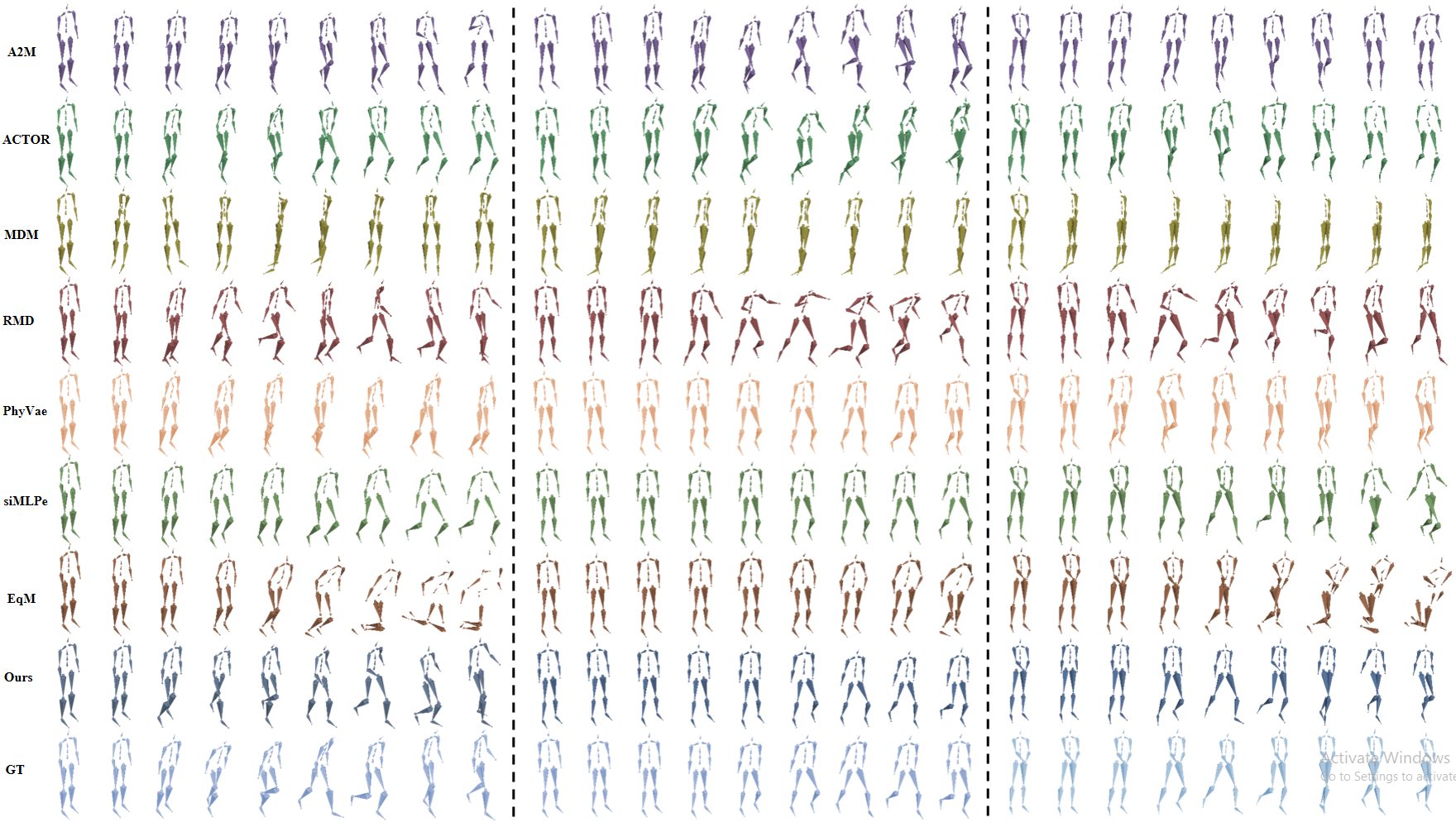}
    \caption{Visual comparison on pushes with different magnitudes. Left: strong, Middle: weak, Right: medium.}
    \label{fig:quality_s_sm}
\end{figure*}
We give the full visual comparison between our methods and the 7 baseline methods in \cref{fig:quality_s_sm}. Overall, our method achieves the best results and is the closest to the ground-truth. Comparatively, MDM and EqM predict visually unreasonable motions with strange poses. A2M, ACTOR, SiMLPe generate visually reasonable snapshots but low-quality motions as well as inaccurate prediction. RMD and PhyVae give more aesthetically pleasing results, but again not high-quality motions and accurate prediction. 

\begin{figure*}
    \centering
    \includegraphics[width=1.0\linewidth]{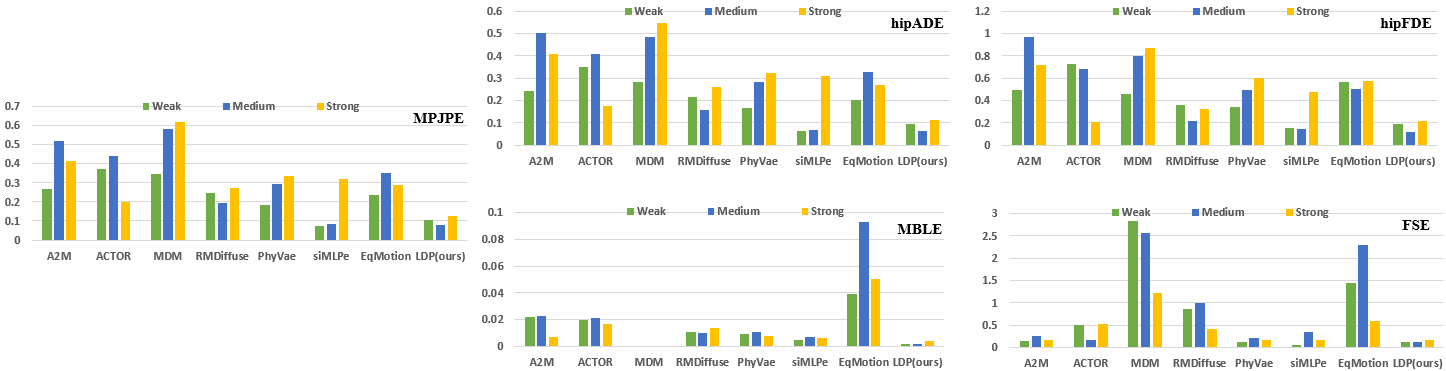}
    \caption{Perturbations with different magnitudes in single-person.}
    \label{fig:inten_s_sm}
\end{figure*}
A more detailed numerical comparison is presented in \cref{fig:inten_s_sm}. Note that the visual comparison is to some extent consistent with the numerical results. MDM and EqM give the worst quality metrics on MBLE and FSE (MBLE for MDM is 0 since it uses a joint-angle-based representation hence no bone-length error). Across all metrics, our model is the best. 

Looking closely, at motion tracking errors, MDM and EqM are not the worst. It suggests that motion tracking metrics and quality metrics evaluate two aspects of the results. This is indeed the case. RMDDiffuse and PhyVae give good motion quality among the baselines and their quality metrics are also good but not necessarily the best. Meanwhile, their tracking metrics are also good but not necessarily the best. A2M can achieve better FSE and sometimes better MBLE than RMDDiffuse and PhyVae, but its motion quality is generally lower. This suggests that there might be a trade-off between motion quality and prediction accuracy among the baselines. But our method achieves the best on both kinds of metrics.

\subsubsection{More Generalization}
\begin{figure*}[tb]
    \centering
    \includegraphics[width=1.0\linewidth]{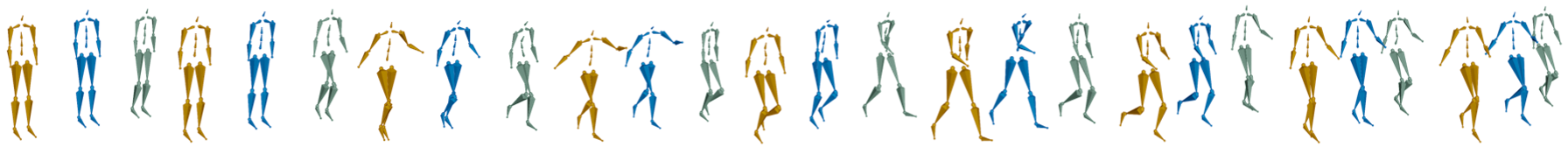}
    \caption{The Generalization to an extra strong push. There are three motions (yellow, blue and green). Blue is the ground truth of a strong push. Yellow is our prediction on the same strong push. Green is an extra strong push.}
    \label{fig:super_sm}
\end{figure*}

We show more generalization experiments in the single-person scenario. We mainly test out-of-distribution push forces in magnitude, timing and duration. In magnitude, we fix the duration of the force to be the same as a strong push but use an extra stronger push that is 37.36\% higher than the strongest push in the dataset. The result is shown in \cref{fig:super_sm}. We can see that the motion pushed by the extra strong push is significantly different from the ground truth and the predicted motion under the strong push. The motion contains earlier foot movements since the initial push is extra strong and it generates a much larger acceleration in the beginning. Also, the upper body is stiffer and has less swing because the balance recovery under an extra strong push tends to require the body to stiffen quickly to prevent the character from falling down and recover balance ultimately.

\begin{figure*}[tb]
    \centering
    \includegraphics[width=1.0\linewidth]{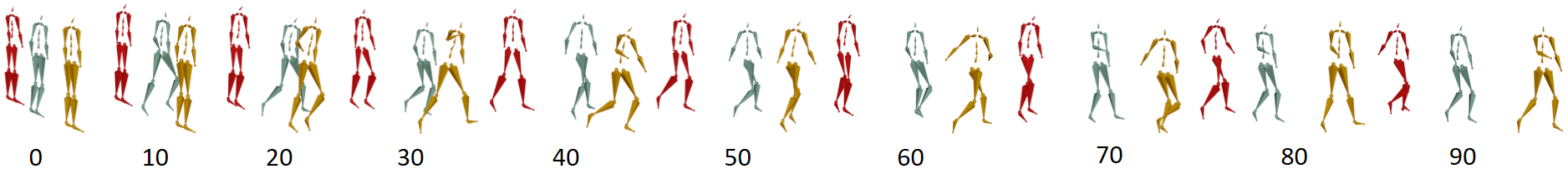}
    \caption{The Generalization to a multi-push scenario. Yellow is the predicted motion under a strong push as in \cref{fig:super_sm}.  Green is the extra strong push in \cref{fig:super_sm}. Red is the three-phase push motion. The numbers indicate the frames.}
    \label{fig:phase_sm}
\end{figure*}
Furthermore, we generalize the push in timing and duration. This time, we apply multiple pushes at different times, as opposed to one push in the beginning as in the data. Note there are not multiple pushes in the data at all. We first apply a weak push, then a medium push at the 15th frame, and finally a strong push at the 50th frame. We show the visual results of this three-phase push in \cref{fig:phase_sm}. One can see that the motion is initially slow and sluggish due to the weak initial push, then gradually intensifies as more pushes are introduced. Under the weak push, the character does not even start to make a step, then it starts to take steps after the medium push at the 15th frame. In the end, large strides need to be made, after the strong push at the 50th frame, to recover balance and counter-balance the accumulated accelerations.

In theory, our model can generalize to other scenarios like slippery surfaces as the friction is learned (Sec 3.1.2 in the main paper). Overall, our model can generalize to out-of-distribution physical disturbances in magnitude, timing and duration.

\subsubsection{Comparison with Full-body Physics-based Models}
\begin{table}[tb]
\small
  \centering
  \begin{tabular}{p{1.5cm}<{\centering} p{1.3cm}<{\centering} p{1.3cm}<{\centering} p{1.3cm}<{\centering}}
    \toprule
    Method & MPJPE & hipADE & hipFDE \\
    \midrule
    PPR & 0.623 & 0.455 & 0.602  \\
   \midrule
   PHC & 0.488 & 0.409 & 0.662  \\
   \midrule
    Ours & 0.097 & 0.086 & 0.171  \\
    \bottomrule

  \end{tabular}
  \caption{Comparison with Full-body Physics-based Baselines.}
  \label{tab:add_bls_sm}
\end{table}

In literature, there is work which also employs body physics for motion imitation under full-body physics-based models~\cite{peng2018deepmimic,gartner2022differentiable,yang2023ppr,luo2023perpetual}.  Although they turn fully/partially observed/user-specified motions into physically valid ones which is different from our task, they could be adapted to our new task. However, they are still intrinsically incapable of learning force interactions in multi-people. So, we could only compare the performance on single-person. To this end, we adapted the latest physics-based models PPR~\cite{yang2023ppr} and PHC~\cite{luo2023perpetual} and compared them with our model in the single-person scenario. Results are shown in \cref{tab:add_bls_sm}. MBLE and FSE are not considered because these two baselines are joint-angle-based and simulation-based.  Overall, Our model still performs best on all metrics. PPR and PHC can generate physically valid motions, but these motions are not necessary accurate prediction.

Compared with the full-body physics models, the Inverted Pendulum Model (IPM) is not fine-grained but has the right granularity for our problem. IPM is a compact yet flexible representation and therefore has been widely used for articulated bodies such as bipedal/quadrupedal robots including humanoids~\cite{hwang2018real}, especially in balance recovery.
Further, simplification is crucial for scalable interaction learning. A full-body model contains 50-100 degrees of freedom (Dofs). Learning from a 4-people scene then involves 200-400 Dofs plus Dofs for interaction forces, which will be extremely unscalable/slow as the learning requires many iterations of forward simulation (for many time steps) and backward propagation. Also, the Dofs will quickly explode in simulation when the number of people increases, \eg our 13-person example. In comparison, one IPM only has 4 Dofs and is much more scalable for both learning and simulation, whose representational capacity has been proven~\cite{kajita20013d,kwon2017momentum}. Also, even with a small model, our model does not overfit, as evidenced by the superb testing results.

Another advantage of using IPMs instead of full-body physics models is the interaction modeling.  We learn interaction forces as potential-energy based forces between two IPMs (Sec 3.1.3), which is flexible and easily learnable.  This is because contact information (position, duration, etc.) is not in the data. Therefore the physics model cannot involve accurate contact modeling even with full-body models, especially when the contact can be frequently established and destroyed in push propagation.


\subsection{Multi-people Results}

\subsubsection{More Comparison}
\begin{figure*}[tb]
    \centering
    \includegraphics[width=0.75\linewidth]
    {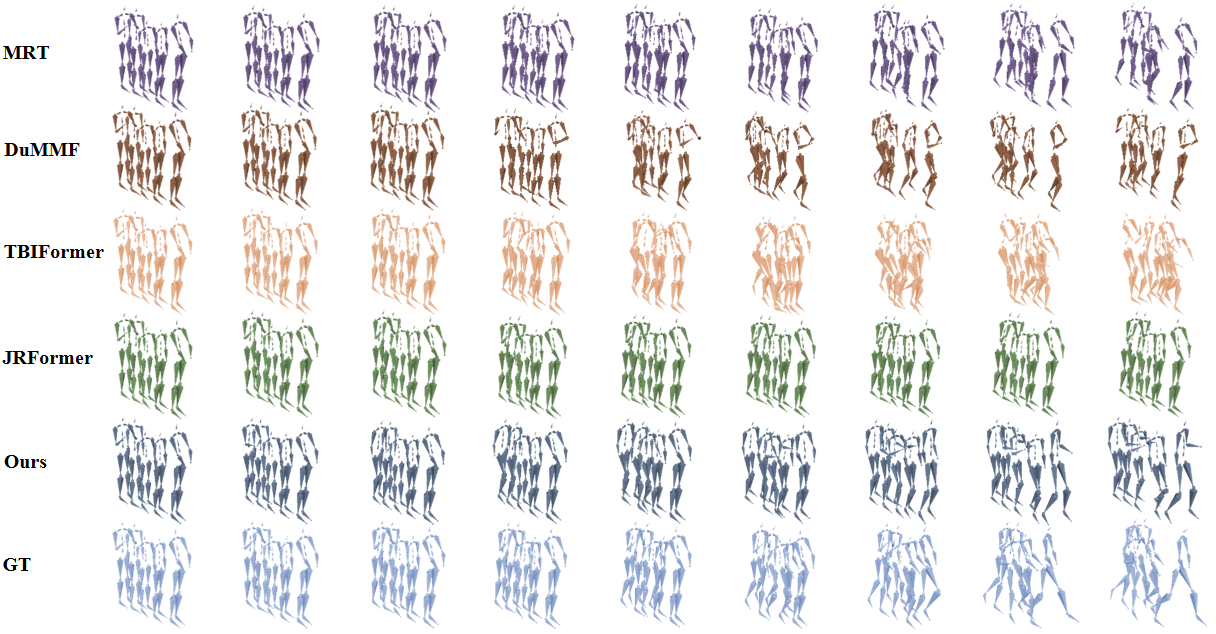}
    \includegraphics[width=0.75\linewidth]{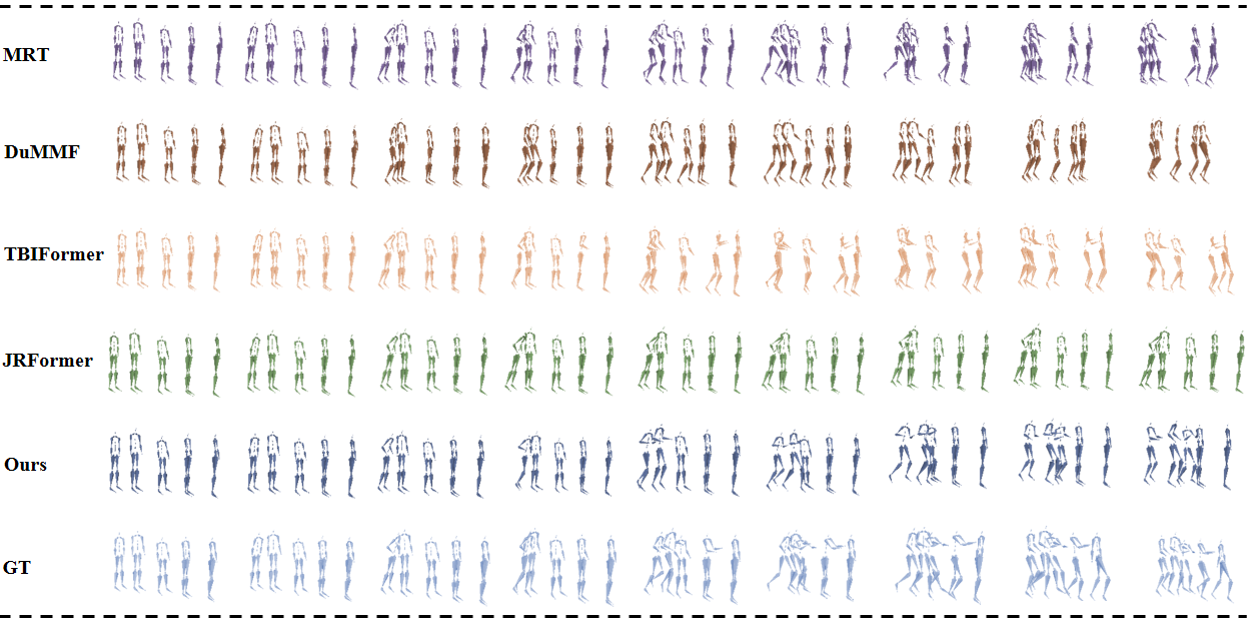}
    \includegraphics[width=0.75\linewidth]{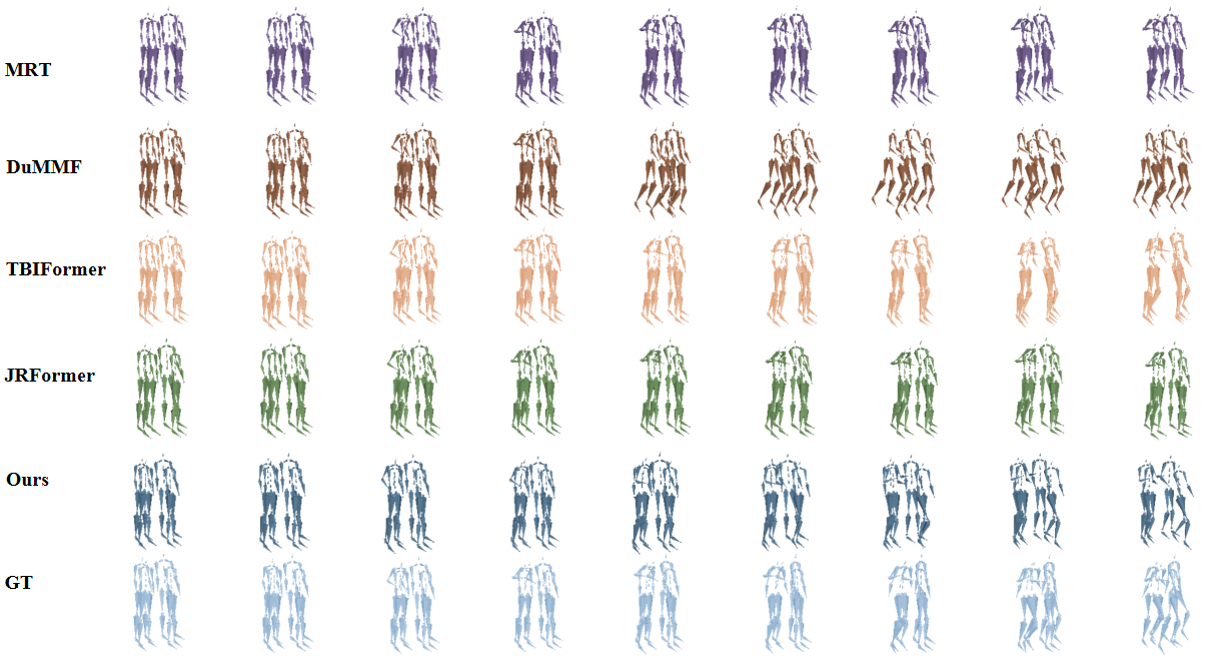}
    \caption{Visual comparison on pushes with different magnitudes and group formations. Top: medium, Middle: strong, Bottom: weak.}
    \label{fig:quality_m_sm}
\end{figure*}

We provide the complete visual comparison between our model with the 4 baseline methods in \cref{fig:quality_m_sm}. Overall, our model obtains the best motion quality and is the closest to the ground truth. DuMMF cannot produce natural movements. JRFormer tends to predict merely subtle motions deviating from the ground truth. MRT and TBIFormer suffer from severe intersections between people for the group formation that is a line. MRT generates serious foot skating for the group formation where people stand in two lines, while TBIFormer performs as well as our model in this formation. Note that all baselines here are given much more information than our model. See the video for a more intuitive comparison.

\begin{figure*}[tb]
    \centering
    \includegraphics[width=1.0\linewidth]{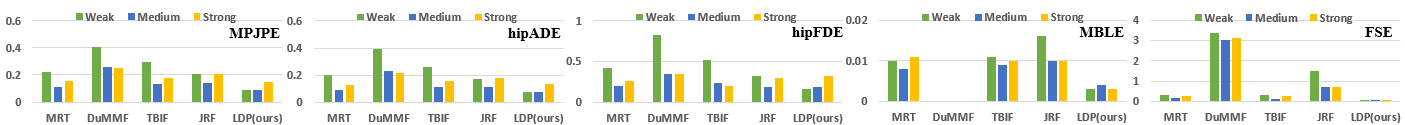}
    \caption{Perturbations with different magnitudes in multi-person.}
    \label{fig:inten_m_sm}
\end{figure*}

Detailed numerical comparison can be found in \cref{fig:inten_m_sm}. DuMMF employs the joint-angle-based representation, resulting in a zero MBLE. Overall, our model achieves or is close to the best performance across metrics and perturbation levels. For all tracking error metrics, our model is much better than baseline methods. This is because only our model can predict the onset and duration of interaction accurately. In motion quality metrics, our model outperforms all baselines across three perturbation levels, meaning that our motion has the best quality.    

\subsubsection{More Generalisation}
\begin{figure*}[tb]
    \centering
    \includegraphics[width=1.0\linewidth]{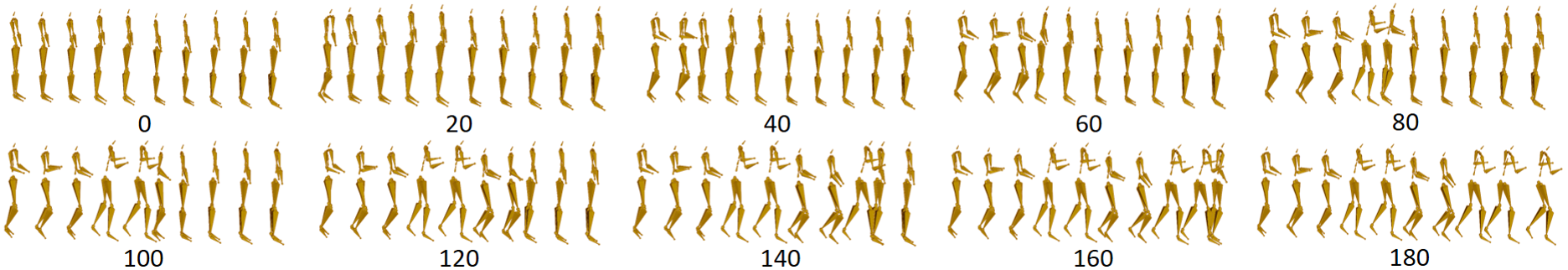}
    \caption{Generalization on Ten people in a line. The first person is pushed by a strong force and we can simulate the force propagation. The number denotes which frame.}
    \label{fig:ten_sm}
\end{figure*}
Other than the 13-people in a diamond formation shown in the main paper, we conduct further generalization experiments. We employ a formation with ten people standing closely in a line, to test whether a strong push can be propagated. Since we explicitly set the distances between people to be very small, we expect a strong push on the first person to be propagated through people all the way to the front, like what is commonly observed in high-density crowds. 

Our prediction results are shown in \cref{fig:ten_sm}. Note this type of scenario is totally out of distribution, in terms of the number of people and the formation (a much longer line).  One can see a clear push propagation starting at the back of the line and then being carried over all the way to the front. This shows not only are the individual motions captured by the model, but the interaction as well as the interaction propagation are also predicted well. 

Furthermore, looking closely at the predicted interaction force between people, we find that the core reason for this push to be propagated, instead of dying down, is that it is intensified by interactions. This is also observed in high-density crowds where a small push can be intensified to cause the ``butterfly effects'' and finally even cause crushes. By turning the parameters in basic forces in the interaction module, it is possible to let the propagation dissipate more quickly. Overall, this shows the great flexibility of our model in capturing complex interactions and interaction propagation. This flexibility could be crucial in crowd simulation in high-density crowds where potential crushes can happen.

\subsection{Data Efficiency}

One core reason for our LDP design is the lack of data. So it is essential to test the data efficiency. Although the original data is already much smaller than existing datasets for human motion prediction, we further reduce the data to 25\% of its original size and repeat the training on single-person and multi-people scenarios.

\begin{table}[tb]
\footnotesize
  \centering
  \begin{tabular}{p{1.5cm} p{1cm} p{0.9cm} p{0.9cm} p{0.8cm} p{0.7cm}}
    \toprule
    Method & MPJPE & hipADE & hipFDE & MBLE & FSE \\
    \midrule
    A2M &0.403 & 0.386 & 0.730 & 0.019 & 0.200  \\
    \midrule
    ACTOR &0.362 & 0.338 & 0.591 & 0.020 & 0.434  \\
    \midrule
    MDM &0.500 & 0.424 & 0.686 & 0 & 2.567  \\
    \midrule
    RMDiffuse & 0.228 & 0.202 & 0.299 & 0.011 & 0.790  \\
    \midrule
    PhyVae & 0.260 & 0.249 & 0.460 & 0.009 & 0.170 \\
    \midrule
    siMLPe & 0.130 & 0.117 & 0.226 & 0.006 & 0.182 \\
    \midrule
    EqMotion & 0.296 & 0.270 & 0.543 & 0.064 & 1.552 \\
    \midrule
    Ours & 0.097 & 0.086 & 0.171 & 0.002 & 0.131 \\
    \bottomrule
    \toprule
    siMLPe\_25\% & 0.203 & 0.189 & 0.411 & 0.009 & 0.650 \\
    \midrule
    Ours\_25\% & 0.207 & 0.190 & 0.267 & 0.009 & 0.211 \\
    \bottomrule

  \end{tabular}
  \caption{Metrics in complete (top) and 25\%  (bottom) training data for single-person.}
  \label{tab:metrics_s_sm}
\end{table}

As shown in \cref{tab:metrics_s_sm}, our model trained on 25\% training data still outperforms all baselines trained 100\% data, except for siMLPe in the single-person scenario. Therefore, we also trained siMLPe on 25\% training data and evaluated it on all metrics for comparison. siMLPe achieves good performance and is slightly better than our model on MPJPE and hipADE on 25\% training data, while our model performs much better on hipFDE and FSE. It's notable that we gave much more information to siMLPe. 

\begin{table*}[th]
\footnotesize
    \centering
    \begin{tabular}{ccccccccc|ccccc}
        \toprule
        Method & A2M & ACTOR & MDM & RMD & PhyVae & siMLPe & EqMotion & Ours\_S & MRT & DuMMF & TBIF & JRF & Ours\_M  \\
        \midrule
        Parameters & 0.45 & 14.78 & 18.10 & 40.96 & 2.72 & 0.02 & 0.64 & 2.67 & 6.98 & 6.54 & 10.26 & 3.70 & 2.94 \\
        \bottomrule
    \end{tabular}
    \caption{Model Size in Single-person (left) and multi-person (right). The unit is M (million). Our\_S means our model for single-person which excludes the differential interaction model. Our\_M is our complete model.}
    \label{tab:model_size_sm}
\end{table*}

\begin{table}[tb]
\footnotesize
  \centering
  \begin{tabular}{p{1.5cm} p{1cm} p{0.9cm} p{0.9cm} p{0.8cm} p{0.7cm}}
    \toprule
    Method & MPJPE & hipADE & hipFDE & MBLE & FSE \\
    \midrule
    MRT & 0.162 & 0.140 & 0.282 & 0.010 & 0.256  \\
    \midrule
    DuMMF & 0.312 & 0.285 & 0.480 & 0 & 3.194  \\
    \midrule
    TBIFormer &0.204 & 0.177 & 0.305 & 0.010 & 0.234  \\
    \midrule
     JRFormer & 0.181 & 0.152 & 0.260 & 0.012 & 0.932 \\
    \midrule
    Ours & 0.106 & 0.092 & 0.218 & 0.003 & 0.069 \\
    \bottomrule
    \toprule
    Ours\_25\% & 0.139 & 0.115 & 0.270 & 0.011 & 0.117 \\ 
    \bottomrule
  \end{tabular}
  \caption{Metrics in complete (top) and 25\%  (bottom) training data for multi-people.}
  \label{tab:metrics_m_s}
\end{table}

One possible reason for the good performance of siMLPe might be its lightweight, as aimed for by its authors. So we also compare the model sizes in \cref{tab:model_size_sm}. It is clear that the lightweight is not the only reason, as other baselines which are smaller than ours cannot achieve good results. We speculate that expressivity especially explicit physics is the key. Further, even siMLPe can achieve good numerical results, its predicted motions are of lower visual quality. More importantly, extending siMLPe to multi-people scenarios is challenging as it cannot learn interactions at all.

Next, we suspect that reduced training data brings more difficulty to the multi-people motion prediction. The results prove us correct, shown in \cref{tab:metrics_m_s}. Our model is still better than all baselines trained on 100\% data, even though the training data for our model is reduced to 25\%. 


The high data efficiency of our model is mainly because the physics model embedded in our model has a low number of learnable parameters, but largely dictates the motion trend. The governing differential equation (Eq. 3 in the main paper) restricts the overall input-output mapping of the whole model and therefore it requires little data to learn. Similar phenomena have been observed in other differentiable physics research~\cite{wang2021sim2sim,yue2023human}.

\section{Additional Experiment Details}
\subsection{Dataset Details}
The new dataset, FZJ Push, records human motions under expected physical perturbations. There are 45 single-person motions and 63 multi-person motions in the dataset. In both scenarios, repeated experiments were conducted on applying unexpected physical pushes with varying magnitude onto a person. In the single-person scenario, this is simply recording reactive motions to push and balance recovery; in multi-people scenario, one person is pushed and this person pushes other people to recover balance so that the push can be propagated among several people.

After discarding redundant frames such as those in waiting, we have 3104 frames and 5614 frames in the single-person and multi-person scenarios, respectively. All pushes are recorded via a pressure sensor Xsensor LX210:50.50.05 on the punching bag. The punching bag was moved manually by the same operator in all experiments. In addition, the pushes are labelled as small, medium and strong.  

\begin{figure}[tb]
  \centering
   \includegraphics[width=0.7\linewidth]{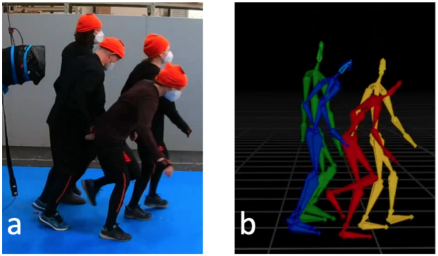}
   \caption{FZJ Push~\cite{feldmann2023forward}. The blue agent was pushed by the punch bag and then he pushed other people.}
   \label{fig:data_sm}
\end{figure}
In the single-person scenario, the dataset involves 4 subjects (S1-S4). \cref{tab:dataset} left shows the number of experiments on each subject under different push magnitudes. We select randomly about 30\% of the data to construct the test set for every subject, while the remaining data is used for training. Finally, the test set and train set have 13 motions and 32 motions, respectively. In the multi-person scenario, the dataset involves 4 group settings. G1 has four people standing in two lines, shown in \cref{fig:data_sm}. G2 has a formation where four people are in a line. G3 and G4 contain 5 people in a line. We give the number of motions in every group setting under different push magnitudes in \cref{tab:dataset} right. We randomly select approximately 20\% of the data in each group setting for the testing set, while the remaining is for training. Eventually, the test set and train set have 14 motions and 49 motions, respectively.         

\begin{table}[tb]
\small
\centering
\begin{tabular}{p{0.6cm} p{0.35cm} p{0.35cm} p{0.35cm} p{0.4cm} | p{0.6cm} p{0.35cm} p{0.35cm} p{0.35cm} p{0.4cm}}
\toprule
Single & Wk & Med & Str & Tot & Group & Wk & Med & Str & Tot \\ 
\midrule
S1 & 3 & 4 & 3 & 10 &G1 & 4 & 4 & 4 & 12\\
\midrule
S2 & 5 & 3 & 4 & 12 &G2 & 6 & 6 & 4 & 16\\
\midrule
S3 & 5 & 4 & 4 & 13 &G3 & 10 & 9 & 6 & 25\\
\midrule
S4 & 3 & 4 & 3 & 10 &G4 & 0 & 10 & 0 & 10 \\
\midrule
Tot & 16 & 15 & 14 & 45 &Tot & 20 & 29 & 14 & 63 \\
\bottomrule
\end{tabular}
\caption{Dataset Details. There are four subjects and four group settings in single-person and multi-people respectively in the dataset. Three push magnitudes (weak, medium and strong) are used.}
\label{tab:dataset}
\end{table}

\subsection{Metrics}
We adopt five metrics commonly used for evaluating motion prediction accuracy and quality as follows. MPJPE (Mean Per Joint Position Error), hipADE (Average Displacement Error at the hip) and hipFDE (Final Displacement Error at the hip) are metrics measuring the tracking errors. MPJPE is the most widely used metric in human motion prediction to evaluate prediction accuracy on every joint. hipADE focuses on the main motion trend, while hipFDE pays attention to the final position of the hip. Moreover, hipADE and hipFDE are strongly relevant to the hip joint which corresponds to the point mass in our IPM. In addition, another two metrics MBLE (Mean Bone Length Error) and FSE (Foot Skating Error) are used to measure the motion quality. We adopt these two metrics to check if our model can produce reasonable poses and motions. 

\begin{itemize}
    \item \textbf{MPJPE}: Mean Per Joint Position Error (MPJPE) is the average $l_2$ distance between predicted positions of joints and their ground truth:
    \begin{align}
        \text{MPJPE}=\frac{1}{TNJ}\sum_{t=1}^T\sum_{n=1}^N\sum_{j=1}^J \lVert X_t^n[j] - \hat{X}_t^n[j] \rVert_2,
    \end{align}
    where $X_t^n[j]$ is the position of the $j$th joint of the $n$th person at frame $t$ and $\hat{X}_t^n[j]$ is its prediction. This metric is used most widely to measure the 3D pose errors. 
    \item \textbf{hipADE}: Average Displacement Error at the hip (hipADE) is the average $l_2$ distance between predicted positions of hip joints and their ground truth:
    \begin{align}
        \text{hipADE} = \frac{1}{TN}\sum_{t=1}^T\sum_{n=1}^N \lVert h_t^n- \hat{h}_t^n \rVert_2,
    \end{align}
    where $h_t^n$ is the hip position of the $n$th person at frame $t$ and $\hat{h}_t^n$ is its prediction. This metric focuses on global errors.
    \item \textbf{hipFDE}: Final Displacement Error at the hip (hipFDE) is the average $l_2$ distance between predicted positions of the hip joints at the last frame in each motion sequence and their ground truth:
        \begin{align}
        \text{hipFDE} = \frac{1}{N}\sum_{n=1}^N \lVert h_T^n- \hat{h}_T^n \rVert_2.
    \end{align}
    \item \textbf{MBLE}: Mean Bone Length Error (MBLE) is the average $l_1$ distance between lengths of predicted bones and their ground truth:
    \begin{align}
        \text{MBLE}=\frac{1}{TNB}\sum_{t=1}^T\sum_{n=1}^N\sum_{b=1}^B \left| X_t^{nb} - \hat{X}_t^{nb} \right|,
    \end{align}
    where $X_t^{nb}$ is the length of $b$th bone of the $n$th person at frame t and $\hat{X}_t^{nb}$ is the corresponding prediction. 
    \item \textbf{FSE}: Foot Skating Error (FSE) is the average of weighted foot velocities for all feet with a height h within a threshold H. The weighted velocity is $v_f (2-2^{h/H})$. 
\end{itemize}

\subsection{Baseline Adaptation}
The task proposed in the main paper is new, so there is no similar work to our best knowledge. For comparison, we adapted 11 state-of-the-art baseline methods in the most relevant areas: motion forecasting, motion generation and motion synthesis. One selection criterion is the availability of the code, to ensure their original implementation is used.

Specifically, we choose A2M~\cite{guo2020action2motion}, ACTOR~\cite{petrovich2021action}, MDM~\cite{tevet2022human}, RMDiffuse~\cite{zhang2023remodiffuse}, PhyVae~\cite{won2022physics}, EqMotion~\cite{xu2023eqmotion}, siMLPe~\cite{guo2023back}, PPR~\cite{yang2023ppr} and PHC~\cite{luo2023perpetual} for the single-person scenario, and MRT~\cite{wang2021multi}, DuMMF~\cite{xu2022stochastic}, TBIFormer~\cite{peng2023trajectory} and JRFormer~\cite{xu2023joint} for the multi-people scenario. We try our best to keep the best performance of these baselines when adapting. The adaptation details are as follows:
\begin{itemize}
    \item \textbf{A2M.} Action2Motion (A2M) is the first work to generate human motions given an action type. We use the push magnitudes (weak, medium and strong) as the action labels (0, 1, 2). The initial pose is applied to kick-start the generation instead of a blank pose filled with 0 in the testing phase.
    \item \textbf{ACTOR.} Action-conditioned Transformer VAE (ACTOR) is another action-to-motion method following A2M. Similar to A2M, the push magnitudes are regarded as the action labels (0, 1, 2). In addition, the initial pose is given when decoding.  
    \item \textbf{MDM.} Motion Diffusion Model (MDM) is one of the first papers employing diffusion models in motion generation. This model can achieve great performance for text-to-motion and action-to-motion. We replace the text input in MDM with the input forces under the text-to-motion setting. Then, the part corresponding to the initial frame in $\hat{x}_0$ is overwritten at each iteration as the MDM does in its motion editing. This is to minimize the change for adaptation. MDM handles motion editing, where if we fix the first frame, the task setting is almost the same as our task. Specifically, motion editing with the initial frame fixed is equivalent to letting the model generate the whole motion given the input signal. 
    \item \textbf{RMDiffuse.} Retrieval-augmented Motion Diffusion model (RMDiffuse) is the state-of-the-art model in motion generation. We adopt its test-to-motion setting and replace the original text input with the input force. Similar to MDM, the part corresponding to the initial frame in $\hat{x}_0$ is overwritten at each iteration during evaluation to ensure the information of the first frame is given.
    \item \textbf{PhyVae}. Physics-based VAE (PhyVae) is the state-of-the-art motion synthesis model. At each step, PhyVae predicts current action $a_t$ given the current input signal $g_t$ and current state $s_t$. Then $a_t$ is fed into a pre-trained network (that can be regarded as a decoder) to predict the next state $s_{t+1}$. The input force at each time step t is regarded as the input signal to synthesize the motion. 
    \item \textbf{siMLPe.} This model is a lightweight network based on MLP but can achieve state-of-the-art performance in single-person motion prediction. For this forecasting approach, it requires as input M frames and predicts N frames. To ensure the comparison is as fair as possible, we provide as input complete information on the input force including magnitude and duration. Specifically, we set M to the maximum duration of the input forces in the single-person scenario. Then, we keep the original ratio between the past and the future frames in the long-term setting in the paper to set N as M/2. M and N values are shown in \cref{tab:baseline}. During testing, given the first M frames, we predict autoregressively to get the complete motion. 
    \item \textbf{PPR and PHC.} These two baselines are state-of-the-art physics-based character animation methods which deal with perturbations. PPR and PHC can synthesize physically valid motions given reference motions. However, reference motions are unavailable during prediction in our new task. Therefore, following the setup in PHC, we use the adapted MDM to generate the reference motions during the test phase. Then these two baselines can generate motions based on the reference motions generated from the adapted MDM.
    \item \textbf{EqMotion, MRT, DuMMF, TBIFormer, JRFormer.} These models fall into human motion forecasting. They have a similar adaptation to that in siMLPe, as they have similar input/output requirements. Details of their settings of input/output frames are shown in \cref{tab:baseline}. EqMotion is the state-of-the-art motion forecasting model for single-person. MRT is a classical multi-person motion prediction method. DuMMF, TBIFormer and JRFormer are state-of-the-art multi-person motion prediction models.  
\end{itemize}

\begin{table}
\footnotesize
  \centering
  \begin{tabular}{p{1.6cm}|p{1cm}|p{1cm}|p{1cm}|p{1cm}}
    \toprule
    \multirow{2}*{Method} &\multicolumn{2}{c}{Original} & 
    \multicolumn{2}{c}{Adaptation}\\
    \cline{2-5}
    & Past & Future & Past & Future \\
    \midrule
    siMLPe &50 & 25 & 12 & 6 \\
    \midrule
     EqMotion & 25 & 25 & 12 & 12 \\
    \midrule
     MRT & 15 & 45 & 20 & 60 \\
    \midrule
     DuMMF & 10 & 25 & 20 & 50 \\
    \midrule
    TBIFormer & 15 & 45 & 20 & 60 \\
    \midrule
    JRFormer & 15 & 45 & 20 & 60 \\
    \bottomrule
  \end{tabular}
  \caption{Adaptation for Motion Prediction Methods. 12 and 20 are the maximum duration of the input forces in the single-person and multi-people scenarios, respectively.}
  \label{tab:baseline}
\end{table}

\subsection{Additional Details of Ablation Study}

Here, we provide more details of the ablation study in the main paper. We conducted the ablation study to evaluate the effectiveness of two important components in our model: the Differentiable IPM and the Skeleton Restoration Model. We have four combinations: with/without IPM, and Full (full-body restoration) / Low-up (first lower body then upper body). 

Our complete model is with IPM and uses a Low-up setting. Without IPM, it means that we only use the Skeleton Restoration Model to predict the next frame, while the two samplers (Upper-sampler and Lower-sampler) have to be dropped as they require the IPM state as input. Therefore, to sample the latent space of the CVAE, we sample the latent variable from a standard Normal distribution during the evaluation phase. The Full/Low-up setting is only within the Skeleton Restoration Model. In Full, we use a Conditional Variational Autoencoder (CVAE) to generate full-body poses directly.  Using the current frame as a condition, we sample the latent space three times and average it. Then both are fed into the decoder to generate the next frame.  In Low-up, we have two CVAEs and we generate the next frame in exactly the same way as in the Full setting, except that we first generate the lower body then the upper body. 

\section{Additional Details of Methodology}
\subsection{Differentiable Inverted Pendulum Model}
Given $I_0$ and $\Dot{I}_0$, we can simulate the IPM motion in time by solving \cref{eq:ipm_mt_sm} repeatedly:
\begin{equation}
  M(I_t, l_{t})\Ddot{I}_t + C(I_t, \Dot{I}_t, l_{t}) + G(I_t, l_{t}) = F^{net}_t 
  \label{eq:ipm_mt_sm}
\end{equation}
where $M\in\mathbb{R}^{4\times4}$, $C\in\mathbb{R}^{4\times1}$ and $G\in\mathbb{R}^{4\times1}$ are the inertia matrix, the Centrifugal/Coriolis matrix, and the external force such as gravity:
\begin{equation}
\resizebox{\hsize}{!}{$
    M_t=\left[\begin{array}{cccc}
         m_c+m_p & 0 & m_p l_t c_{\theta_t} & 0  \\
         0 &  m_c+m_p & m_p l_t s_{\theta_t}s_{\phi_t} & - m_p l_t c_{\theta_t}c_{\phi_t} \\
         m_p l_t c_{\theta_t} & m_p l_t s_{\theta_t}s_{\phi_t} & m_p l_t^2 & 0 \\
         0 & - m_p l_t c_{\theta_t}c_{\phi_t} & 0 & m_p l_t^2 c_{\theta_t}^2
    \end{array}   \right] $} \nonumber
\end{equation}
\begin{equation}
\resizebox{\hsize}{!}{$
    \begin{array}{cc}
      C_t=\left[\begin{array}{c}
        -m_p l_t s_{\theta_t} \Dot{\theta}_t^2 \\
        m_p l_t (2 s_{\theta_t} c_{\phi_t}\Dot{\theta}_t \Dot{\phi}_t + c_{\theta_t} s_{\phi_t} (\Dot{\theta}_t^2 + \Dot{\phi}_t^2)) \\
        m_p l_t^2 s_{\theta_t} c_{\phi_t} \Dot{\phi}_t^2 \\
        -2 m_p l_t^2 s_{\theta_t} c_{\theta_t} \Dot{\theta}_t \Dot{\phi}_t
    \end{array}\right]   
      &
      G_t=\left[\begin{array}{c}
      0 \\
      0 \\
      -m_p g l_t s_{\theta_t} c_{\phi_t} \\
      -m_p g l_t c_{\theta_t} s_{\phi_t}
      \end{array} \right]
    \end{array}. 
    $} \nonumber
\end{equation}
Here, $m_c$ and $m_p$ are the mass of the cart and the pendulum, respectively. $c_{\theta_t}$ and $s_{\theta_t}$ denote $\cos{\theta_t}$ and $\sin{\theta_t}$, while $c_{\phi_t}$ and $s_{\phi_t}$ represent $\cos{\phi_t}$ and $\sin{\phi_t}$. We set $m_c$ and $m_p$ as 0.1$M$ and 0.9$M$ respectively where $M$ is the total mass of a person.   
Unlike the standard IPM, we allow the rod length to change with time.  Given the net force $F^{net}_t\in\mathbb{R}^4$ and the rod length $l_{t}$, we can solve \cref{eq:ipm_mt_sm} for the next state $I_{t+1}$ via a semi-implicit scheme:
\begin{align}
\Dot{I}_{t+1}  = \Dot{I}_t + \triangle t \Ddot{I}_t, \quad
I_{t+1}  = I_t + \triangle t \Dot{I}_{t+1},    \nonumber
\end{align}
where $\triangle t$ is the time step. We have elaborated on the prediction of  $F^{net}_t$ and $l_{t}$ in the main paper. Then we have the following equation:
\begin{equation}
\resizebox{\hsize}{!}{$
    I_T - I_0  = \int_0^{T} \Dot{I}_t \mathrm{d}t = \int_0^{T} \int M_t^{-1}(F_t^{net} - C_t - G_t) \mathrm{d}t\mathrm{d}t, $} \nonumber
\end{equation}
given the initial condition $I_0$ and $\Dot{I}_0$ and the final station $I_T$. The prediction of $F^{net}_t$ is based on the neural networks and other differentiable operations such as PD control and repulsive potential energy. The prediction of the rod length $l_t$ is from a neural network. Finally, the semi-implicit scheme for updating $I_t$ only includes simple differentiable arithmetic. Therefore, our complete IPM is differentiable for both single-person and multi-person scenarios.

\textbf{Single-person Prediction.} The hyper-parameters $K_p$ and $K_d$ in the PD control are [30, 30, 1500, 1500] and [4, 4, 200, 200], respectively.  We use an LSTM with the size 256 to predict $F_t^{self-nn}$. The MLP predicting the rod length has hidden size [128, 128].

\textbf{Differential Interaction Model.} if $|r_{t,nj}| < r_{neigh}$, the $j$th person is the neighbor of the $n$th person at time $t$ \ie $j \in \Omega_{t,n}$, where $r_{neigh} = 0.5$. We use an MLP with 2 hidden layers [512, 512] to predict $F_{t, nj}^{inta-nn}$. The hyper-parameters $u$ and $\sigma$ in the repulsive potential energy function for calculating $F_{nj}^{bs-xy}$ are 150 and 0.5, respectively. Then, we elaborate the $F_{nj}^{bs-\theta\phi} = [F_{nj}^{bs-\theta}, F_{nj}^{bs-\phi}]^{\mathbf{T}}$. We give details for $F_{nj}^{bs-\theta}$, where the same principle also applies to $F_{nj}^{bs-\phi}$. The magnitude of $F_{nj}^{bs-\theta}$ is a constant $k_{\theta} = 100$ ($k_{\phi} = 50$), while its direction is based on the $\theta_n$ and $\theta_j$ of IPM states of $n$ and $j$. We categorize $\theta$ into three groups: positive, zero and negative. For two IPMs, this produces a total of 9 possible situations. Then we need to decide their relative position. Taking the $n$th person as the person in interest, if its relative position with respect to a neighbor $j$ along the x-axis is positive \ie $x_{nj} =  x_n - x_j > 0$, we label it as BE, otherwise BA. We show the directions of the force for all 9 possible situations in \cref{tab:inta_sm}, where 1 denotes the interaction force is positive, 0 denotes no interaction forces, and -1 denotes the negative direction.    

\begin{table}[tb]
\footnotesize
    \centering
    \begin{tabular}{c|p{0.5cm}|p{0.5cm}|p{0.5cm}|c|p{0.5cm}|p{0.5cm}|p{0.5cm}}
    \toprule
    \diagbox[width=1cm]{$\theta_n$}{$\theta_j$}   & Pos & Zero & Neg &  \diagbox[width=1cm]{$\phi_n$}{$\phi_j$} & Pos & Zero & Neg \\
    \midrule
    Pos & 1/-1 & 0/-1 & 0/-1 & Pos & -1/1 & -1/0 & -1/0   \\
    \midrule
    Zero& 1/0 & 0/0 & 0/-1 & Zero& 0/1 & 0/0 & -1/0  \\
    \midrule
    Neg & 1/0 & 1/0 & 1/-1 & Neg & 0/1 & 0/1 & -1/1 \\
    \bottomrule
    \end{tabular}
    \caption{Basic Interaction Force on Angles. X/X is BE/BA.}
    \label{tab:inta_sm}
\end{table}
\begin{figure}[tb]
    \centering
    \includegraphics[width=1.0\linewidth]{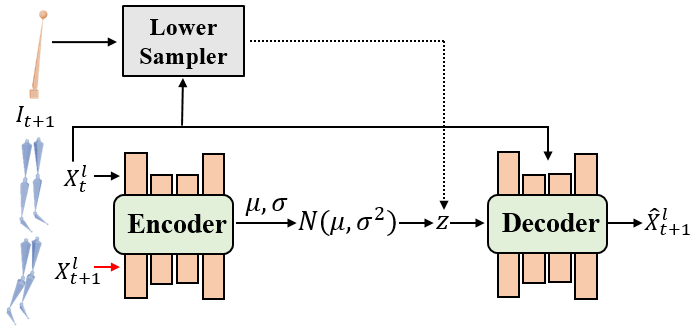}
    \caption{The Architecture of the CVAE-Lower. During training, the current lower-body pose $X_t^l$ as the condition, and the next lower-body pose $X_{t+1}^l$ are fed into the encoder to predict the distribution of the latent variable z. Then the decoder predicts the next pose $\hat{X}_{t+1}^l$ from the sampled variable z and the condition. The red connection is only used in training. During inference, we use the lower sampler to sample the latent variable $z$ to predict motion.}
    \label{fig:cvae_l_sm}
\end{figure}
\begin{figure}[tb]
    \centering
    \includegraphics[width=1.0\linewidth]{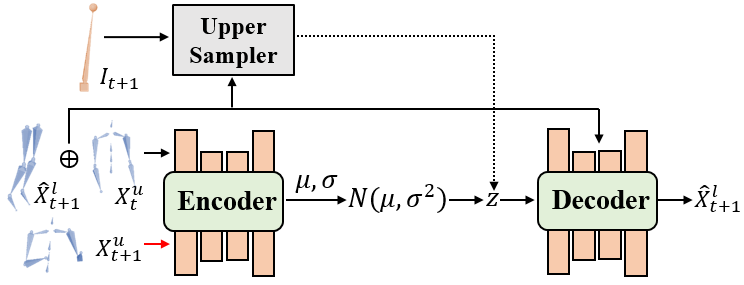}
    \caption{The Architecture of the CVAE-Upper. The condition $X_t^u$ and $\hat{X}_{t+1}^l$, together with the next pose $X_{t+1}^u$ are fed into the encoder to predict the distribution of the latent variable z. Then the decoder predicts the next pose $\hat{X}_{t+1}^u$ from the sampled variable z and the condition. The red connection is only used in training. During inference, we use the upper sampler to sample the latent variable $z$ to predict motion.}
    \label{fig:cvae_u_sm}
\end{figure}
\subsection{Skeleton Restoration Model}
\textbf{Lower Body Restoration.} We follow~\cite{ling2020character} to construct the CVAE-Lower as shown in \cref{fig:cvae_l_sm}. The encoder is an MLP with two hidden layers of 256 dimensions, with an ELU layer following each hidden layer. The dimension of the latent variable z is 64. A mixture-of-expert architecture is employed for the decoder, including 4 expert networks and a gating network. The input to the gate network and the expert networks are the latent variable z combined with the current lower-body pose $X_t^l$, while the output of the expert network is the next pose $X_{t+1}^l$.

Similar to the encoder, the gate network is an MLP with two 64D hidden layers followed by ELU activations. Each expert network has the same structure as the encoder except for the input layer and the output layer. 

During testing, we use the Lower Sampler to sample the latent variable z given the current lower-body pose $X_t^l$ and the predicted IPM state $I_{t+1}$. The Lower sampler has the same structure as that of the encoder except for the input layer. 

\textbf{Upper Body Restoration.} The CVAE-Upper has the same architecture as the CVAE-Lower except for the condition $X_t^u$ and $\hat{X}_{t+1}^l$ as shown in \cref{fig:cvae_u_sm}. Similarly, the upper sampler has the same structure as the encoder of CVAE-Upper except for the input layer. 
Although the upper body is not explicitly physically constrained, it is implicitly constrained by the IPM motion which is physically based.

\textbf{State Representation.} In the skeleton restoration model, we adopt pose representations~\cite{ling2020character, zhang2018mode} for the full-body pose. Specifically, we use a vector including positions, rotations, and velocities to represent the pose $X_t$. $X_t^l$ and $X_t^u$ take the corresponding lower or upper part in the $X_t$. Furthermore, we use a 15D vector $[x_t, y_t, \theta_t, \phi_t, e_t, l_t, \Dot{x_t}, \Dot{y_t}, \Dot{\theta_t}, \Dot{\phi_t}, \Dot{e_t}]$ for $I_t$ and input the vector into the sampler, where $e_t$ is the position of the end of the rod corresponding to the hip joint and $l_t$ is the rod length. 

\subsection{Training}
There are several components in our model. An end-to-end training but could lead to suboptimal local minima. Therefore, we employ pre-training to initialize individual components and also use auxiliary losses in addition to the main loss introduced in the main paper.

We train the IPM first. Then, we train the CVAE-Lower. Next, we train the lower sampler network based on the trained CVAE-Lower. Similarly, we train the CVAE-Upper first then the upper sampler network. 

We train the differentiable IPM model by using the 0-order and 1-order information as shown in \cref{eq:l_ipm_sm}, where $\lambda$ is a weight parameter. We minimize the angular velocity $\Dot{\phi}$ instead of penalizing its $l_1$ norm as we do in other dimensions. This is because the angular velocity should always be smooth when recovering balances so that smoothing leads to better results than minimizing the $l_1$ norm.       
\begin{align}
    \label{eq:l_ipm_sm}
    L_{ipm} = & \frac{1}{T} \sum_{t=1}^T \{ |\hat{x}_t - x_t| + |\hat{y}_t - y_t| + |\hat{\theta}_t - \theta_t| + |\hat{\phi}_t - \phi_t| \nonumber \\
    +& |\hat{\Dot{x}}_t - \Dot{x}_t| + |\hat{\Dot{y}}_t - \Dot{y}_t| + |\hat{\Dot{\theta}}_t - \Dot{\theta}_t| + \lambda|\hat{\Dot{\phi}}_t| \}
\end{align}

We follow \cite{ling2020character} to train the CVAE-Lower in our skeleton restoration model. Then we train the Lower sampler network based on the trained CVAE-Lower. The encoder of the CVAE-Lower and the lower Sampler both output the Gaussian distribution parameters $[\mu, \sigma]$ for the latent variables $z$. We train the lower sampler by using the loss function in \cref{eq:l_skel_sm} to let the outputs of the Lower Sampler be close to those of the encoder, and we ensure that the restored poses have low FSE.   
\begin{align}
    \label{eq:l_skel_sm}
    L_{skel} = \| \hat{z}_{\mu} - z_{\mu} \|^2 + \| \hat{z}_{\sigma} - z_{\sigma} \|^2 + \text{FSE}(\hat{X}_t^{l}, X_t^{l})
\end{align}
We train the CVAE-Upper and the Upper sampler network in the same way except that the FSE in the loss function \cref{eq:l_skel_sm} is ignored. 

After initialization, the whole network can be trained as a whole. We use the Adam optimizer for all training. The learning rates for training the differentiable IPM and two samplers are 3e-4 and 1e-4, respectively. When training the CVAEs, a linear schedule is used to adjust the learning rate from 1e-4 to 1e-7, and we set the weight of the KL loss as 0.005 to encourage high reconstruction quality. The whole training takes about 15 hours on a single GeForce RTX 2080 Ti, but can be automated. The inference takes approximately 0.19 sec/frame in our 13-person experiment.
{
    \small
    \bibliographystyle{ieeenat_fullname}
    \bibliography{main}
}


\end{document}